\documentclass[sigconf,table]{acmart}

\usepackage[english]{babel}
\usepackage{type1ec}
\usepackage[utf8]{inputenc}
\usepackage[T1]{fontenc}
\usepackage{subcaption,graphicx}
\usepackage{pbox}
\usepackage{makecell}

\usepackage{amssymb}
\usepackage{amsthm}
\usepackage{multirow}
\usepackage{fancyvrb}
\usepackage{algpseudocode}
\usepackage{algorithm}
\usepackage{algpseudocode}
\usepackage{subcaption}
\usepackage{mathtools}
\usepackage{qtree}
\usepackage{alltt}
\usepackage{lipsum}  
\usepackage{makecell}
\usepackage{tablefootnote}
\usepackage{bibentry}
\usepackage{geometry}
\usepackage{booktabs} 

\usepackage{color, colortbl}

\definecolor{officegreen}{rgb}{0.0, 0.5, 0.0}
\definecolor{byzantium}{rgb}{0.44, 0.16, 0.39}
\definecolor{darkturquoise}{rgb}{0.0, 0.81, 0.82}

\setcopyright{rightsretained}

\acmDOI{10.1145/nnnnnnn.nnnnnnn}

\acmISBN{978-x-xxxx-xxxx-x/YY/MM}

\acmConference[GECCO '20]{the Genetic and Evolutionary Computation Conference 2020}{July 8--12, 2020}{Cancun, Mexico}
\acmYear{2020}
\copyrightyear{2020}

\acmPrice{15.00}

\renewcommand\footnotetextcopyrightpermission[1]{} 
\begin{document}
\title{A Robust Experimental Evaluation of Automated Multi-Label Classification Methods}
\subtitle{Supplementary Material: Proposed \emph{Search Spaces}}

\author{Alex G. C. de S\'a}
\affiliation{%
  \institution{Computer Science Department \\ Universidade~Federal~de~Minas~Gerais}
  \city{Belo Horizonte, MG} 
  \state{Brazil} 
}
\email{alexgcsa@dcc.ufmg.br}

\author{Cristiano G. Pimenta}
\affiliation{%
  \institution{Computer Science Department \\ Universidade~Federal~de~Minas~Gerais}
  \city{Belo Horizonte, MG} 
  \state{Brazil} 
}
\email{cgpimenta@dcc.ufmg.br}

\author{Gisele L. Pappa}
\affiliation{%
  \institution{Computer Science Department \\ Universidade~Federal~de~Minas~Gerais}
  \city{Belo Horizonte, MG} 
  \state{Brazil} 
}
\email{glpappa@dcc.ufmg.br}

\author{Alex A. Freitas}
\affiliation{
  \institution{School of Computing \\ University of Kent}
  \city{Canterbury,~Kent} 
  \state{United Kingdom} 
}
\email{a.a.freitas@kent.ac.uk}

\renewcommand{\shortauthors}{}

\begin{abstract}
This supplementary material aims to describe the proposed multi-label classification (MLC) \emph{search spaces} based on the MEKA and WEKA softwares. First, we overview 26 MLC algorithms and meta-algorithms in MEKA, presenting their main characteristics, such as hyper-parameters, dependencies and constraints. Second, we review 28 single-label classification (SLC) algorithms, preprocessing algorithms and meta-algorithms in the WEKA software. These SLC algorithms were also studied because they are part of the proposed MLC \emph{search spaces}. Fundamentally, this occurs due to the problem transformation nature of several MLC algorithms used in this work. These algorithms transform an MLC problem into one or several SLC problems in the first place and solve them with SLC model(s) in a next step. Therefore, understanding their main characteristics is crucial to this work. Finally, we present a formal description of the \emph{search spaces} by proposing a context-free grammar that encompasses the 54 learning algorithms. This grammar basically comprehends the possible combinations, the constraints and dependencies among the learning algorithms.
\end{abstract}

\maketitle

\section{Studying the search space of multi-label classification algorithms}
\label{searchspacestudy}

We perform a study of \textbf{26} multi-label and meta multi-label classification algorithms from the MEKA software~\citep{Read2016}, which are described in the following two sections.  

Table \ref{mlc:tab1} lists these 26 MLC algorithms and present their association to the  different versions of the designed \emph{search spaces}, showing their names, names' acronyms and their respective types. Table \ref{mlc:tab1} also indicates if the MLC algorithm in the row belongs or not (`Y' for yes and `N' for no) to the respective \emph{search space} in the column (i.e., if the MLC \emph{search space} comprehends that algorithm), and how many hyper-parameters (\#HP) it encompasses (when it is used). 

\begin{table*}[!htbp]

\begin{center}
 \caption{Overview of the employed multi-label classification (MLC) algorithms from the MEKA software$^*$.}
 \label{mlc:tab1}
\begingroup

\begin{tabular}{|l|l|l|l||c|c||c|c||c|c|}
\cline{5-10}
\multicolumn{4}{l}{}                                                          & \multicolumn{2}{|c|}{\textbf{Small}} & \multicolumn{2}{|c|}{\textbf{Medium}} & \multicolumn{2}{|c|}{\textbf{Large}} \\\hline
\textbf{id} & \textbf{Algorithm Name}                                      & \textbf{Acronym}      & \textbf{Type}     & \textbf{Used?}          & \textbf{\#HP}          & \textbf{Used?}            & \textbf{\#HP}           & \textbf{Used?}           & \textbf{\#HP}           \\\hline             
 1 &   \hyperref[mlc:bpnn]{Back Propagation Neural Network}                & ML-BPNN      & AA       & Y             & 4               & Y               & 4                & Y              & 4                \\\hline

 2 & \hyperref[mlc:br]{Binary Relevance}                                & BR           & PT       & Y             & 0               & Y               & 0                & Y              & 0                \\\hline
 3 &  \hyperref[mlc:cc]{Classifier Chain}                               & CC           & PT       & Y             & 0               & Y               & 0                & Y              & 0                \\\hline
 4 &  \hyperref[mlc:lc]{Label Powerset}                                 & LP           & PT       & Y             & 0               & Y               & 0                & Y              & 0                \\\hline
 5 &  \hyperref[mlc:rakel]{Random K-Label Pruned Sets}                  & RAkEL        & PT       & Y             & 4               & Y               & 4                & Y              & 4                \\\hline
  \hline
  
 6 &  \hyperref[mlc:bcc]{Bayesian Classifier Chains}                    & BCC          & PT       & N             & -               & Y               & 1                & Y              & 1                \\\hline  
 7 &  \hyperref[mlc:brq]{Binary Relevance -- Quick Version}             & BRq          & PT       & N             & -               & Y               & 1                & Y              & 1               \\\hline
 8 &  \hyperref[mlc:ccq]{Classifier Chain -- Quick Version}             & CCq          & PT       & N             & -               & Y               & 1                & Y              & 1                \\\hline
 9 &  \hyperref[mlc:fw]{Four-Class Pairwise Classification}             & FW           & PT       & N             & -               & Y               & 0                & Y              & 0               \\\hline
 10 &  \hyperref[mlc:mcc]{Monte-Carlo Classifier Chains}                & MCC          & PT       & N             & -               & Y               & 3                & Y              & 3                \\\hline
 11 &  \hyperref[mlc:pcc]{Probabilistic Classifier Chains}              & PCC          & PT       & N             & -               & Y               & 0                & Y              & 0               \\\hline    
 12 &  \hyperref[mlc:ps]{Pruned Sets}                                   & PS           & PT       & N             & -               & Y               & 2                & Y              & 2                \\\hline
 13 &  \hyperref[mlc:pst]{Pruned Sets with Threshold}                   & PSt          & PT       & N             & -               & Y               & 2                & Y              & 2   
  \\\hline  
 14 &  \hyperref[mlc:rakeld]{Random K-Label  Disjoint Pruned Sets}      & RAkELd       & PT       & N             & -               & Y               & 3                & Y              & 3                \\\hline  
 15 & \hyperref[mlc:rt]{Ranking and Threshold}                          & RT           & PT       & N             & -               & Y               & 0                & Y              & 0                \\\hline    
  \hline
 16 & \hyperref[mlc:ct]{Classifier Trellis}                             & CT           & PT       & N             & -               & N               & -                & Y              & 6                \\\hline   
 17 & \hyperref[mlc:cdn]{Conditional Dependency Networks}               & CDN          & PT       & N             & -               & N               & -                & Y              & 2               \\\hline
 18 & \hyperref[mlc:cdt]{Conditional Dependency Trellis}                & CDT          & PT       & N             & -               & N               & -                & Y              & 5                \\\hline  
 19 & \hyperref[mlc:pmcc]{Population of MCC}                            & PMCC         & PT       & N             & -               & N               & -                & Y              & 6                \\\hline

 20 & \hyperref[mlc:bagging]{Bagging of Multi-Label Classifiers}             & BaggingML    & Meta-MLC & N             & -               & N               & -                & Y              & 1                \\\hline  
 21 & \hyperref[mlc:baggingdup]{Bagging of Multi-Label Classifiers (Duplicate)} & BaggingMLDup & Meta-MLC & N             & -               & N               & -                & Y              & 2                \\\hline 
 22 & \hyperref[mlc:cm]{Classification Maximization}                    & CM           & Meta-MLC & N             & -               & N               & -                & Y              & 1                \\\hline

 23 & \hyperref[mlc:ensemble]{Ensemble of Multi-Label classifiers}           & EnsembleML   & Meta-MLC & N             & -               & N               & -                & Y              & 2                \\\hline 
 24 & \hyperref[mlc:em]{Expectation Maximization}                      & EM           & Meta-MLC & N             & -               & N               & -                & Y              & 1               \\\hline 
 25 & \hyperref[mlc:rsml]{Random Subspace Multi-Label}                   & RSML         & Meta-MLC & N             & -               & N               & -                & Y              & 3                \\\hline
 26 & \hyperref[mlc:sm]{Subset Mapper}                                 & SM           & Meta-MLC & N             & -               & N               & -                & Y              & 0  \\ \hline 
\end{tabular}                                                                                                        
\endgroup
\begin{tabular}{l} 
$^*$All algorithm names are clickable links to their respective part in this supplementary material. \hspace{16cm} \text{ }
\end{tabular}
\end{center}

\end{table*}

It is also important to mention that the algorithms in (this version of) MEKA can define a \textbf{threshold} to perform the classification using the model's confidence outputs (typically, class probabilities). 
For the general multi-label context, it is in general better to optimize the threshold than simply using an arbitrary threshold of 0.5~\citep{Fan2007,Read2011}.
This parameter ($pred\_tshd$) [\emph{-threshold}] could take the following values:
\begin{itemize}
 \item Proportional cut method by instance (PCut1) \citep{Read2010}: 
 It takes into account the label cardinality of the dataset, which is simply the average number of labels associated with each instance of this dataset. Thus, 
 PCcut1 automatically calibrates the prediction confidence threshold, by
 minimizing the difference between the label cardinality of the training set and the label cardinality obtained with a given set of predicted labels -- where the latter
 set is determined by the threshold value. 
 This does not require access to the true predictions in the test set.
 \item Proportional cut method by label (PCutL): It is used to calibrate the prediction confidence threshold the same way as PCut1, but for each label individually.
 \item The threshold could also take a unique real value between zero (0.0) and one~(1.0) for all instances being classified. Formally,  the threshold can also be defined by the following interval: 
 $\{threshold \in \mathbb{R} \text{ } | \text{ } 0.0 < threshold < 1.0 \}$.
\end{itemize}
 \textbf{Default value: }PCut1.



\section{Search Space -- Traditional MLC Algorithms}
\label{mlc}

This section covers the main traditional MLC algorithms in the MEKA software \cite{Read2016}. 

\subsection{Binary Relevance}\label{mlc:br}
\label{br-method}

The standard binary relevance (BR) algorithm \citep{Tsoumakas2010}. It creates a binary classification problem for each label and learns a model for each label individually.
\textbf{Parameters:}
\begin{itemize}
 \item \underline{Base classifier (\emph{bc})[-W] }: It can be any classifier from WEKA.
\end{itemize}

There are no dependencies/constraints in BR.


\subsection{The `quick' version of Binary Relevance}\label{mlc:brq}

The quick version of BR (BRq) \citep{Read2011} is a version of BR which is able to downsample the number of training instances across the binary models.
It is intended for use in an ensemble (but  it works in a standalone fashion as well).
 
\textbf{Parameters:}
\begin{itemize}
 \item \underline{Base classifier (\emph{bc})[-W] }: It can be any classifier from WEKA.
 
 \item \underline{Down-sample ratio (\emph{dsr})[-P]}: It is a ratio used to reduce the number of instances across the binary models.
 Low values mean more removals and high values mean less removals, as BRq uses the following formula $(1 - dsr) * number\_of\_instances$ to calculate
 the number of instances to remove. This parameter is constrained by the interval: \\ $\{dsr \in \mathbb{R} \text{ } | \text{ } 0.2 \leq dsr \leq 0.8 \}$. \\
 There is no explanation about this parameter in the original paper and in other papers in the multi-label classification literature.
 The justification -- about the used interval -- is that we would like to have (at least) 20\% of the 
 instances from the original data to construct the model (otherwise, the algorithm may not have sufficient instances to build the model).
 Additionally, we would like to have (at most) 80\% of the instances from the original data in order to learn the classifier (otherwise, the 
 algorithm would be very similar to BR).
 \\ \textbf{Default value: }0.75.

\end{itemize}

There are no dependencies/constraints between the parameters of BRq.


\subsection{Classifier Chain}\label{mlc:cc}

The classifier chain (CC) algorithm \citep{Read2011} is also similar to BR, but the label outputs predicted by a classifier become new inputs for the next classifiers in the chain.
It uses a single random order of labels in the chain.

\textbf{Parameters:}
\begin{itemize}
 \item \underline{Base classifier (\emph{bc})[-W] }: It can be any classifier from WEKA.
 
\end{itemize}

There are no dependencies/constraints in CC.

\subsection{The `quick' version of Classifier Chains}\label{mlc:ccq}

The quick version of CC (CCq) \citep{Read2011} is a version of CC which is able to down-sample the number of training instances across the binary models.
It is also intended for use in an ensemble (but  it works in a standalone fashion as well).
\textbf{Parameters:}
\begin{itemize}
 \item \underline{Base classifier (\emph{bc})[-W] }: It can be any classifier from WEKA.
 
 \item \underline{Down-sample ratio (\emph{dsr})[-P]}: It is a ratio used to reduce the number of instances across the binary models.
 Low values mean more removals and high values mean less removals, as CCq uses the following formula $(1 - dsr) * number\_of\_instances$ to calculate
 the number of instances to remove. This parameter is constrained by the interval: \\ $\{dsr \in \mathbb{R} \text{ } | \text{ } 0.2 \leq dsr \leq 0.8 \}$. \\
 There is no explanation about this parameter in the original paper and in other papers in the multi-label classification literature.
 The justification -- about the used interval -- is that we would like to have (at least) 20\% of the 
 instances from the original data to construct the model (otherwise, the algorithm may not have sufficient instances to build the model).
 Additionally, we would like to have (at most) 80\% of the instances from the original data in order to learn the classifier (otherwise, the 
 algorithm would be very similar to CC).
 \\ \textbf{Default value: }0.75.

\end{itemize}

There are no dependencies/constraints between the parameters of CCq.

\subsection{Bayesian Classifier Chain}\label{mlc:bcc}


The Bayesian classifier chain (BCC) algorithm \citep{Zaragoza2011} creates a maximum spanning tree based on marginal dependencies, define a Bayesian network from it, and then employ a classifier chain (CC) using the order of the labels found in the Bayesian network model. The original paper used Na\"ive Bayes as a base classifier, but other types of classifiers can be used. \textbf{Parameters:}
\begin{itemize}
 \item \underline{Base classifier (\emph{bc})[-W] }: It can be any classifier from WEKA.
 \item \underline{Dependency type (\emph{dp})[-X]}: The way to measure and find the dependencies. It may take ten categorical values:
  1. C (co-occurrence counts);   2. I (mutual information); 3. Ib (mutual information using binary approximation);
  4. Ibf (Mutual information using fast binary approximation); 5. H (Conditional information);
  6. Hbf ( Conditional information using fast binary approximation); 7. X (Chi-squared);
  8. F (Frequencies); 9. No label dependence; 10. L (The ``LEAD'' method for finding conditional dependence).
   \\ \textbf{Default value: }Ibf.
\end{itemize}

There are no dependencies/constraints between the parameters of BCC.

\subsection{(Bayes Optimal) Probabilistic Classifier Chain}\label{mlc:pcc}

The probabilistic classifier chain (PCC) algorithm \citep{Dembczynski2010} acts exactly like CC at training time, but explores all possible paths as inference at test time (hence, ``Bayes optimal'').
\textbf{Parameters:}
\begin{itemize}
 \item \underline{Base classifier (\emph{bc})[-W]}: It can be any classifier from WEKA.
\end{itemize}

\textbf{Dependencies/Constraints:}
\begin{enumerate}
 \item PCC has poor scalability, i.e.,  it is very slow when the number of labels is greater than a certain threshold. 
In the PCC's original paper \citep{Dembczynski2010}, it is said that this threshold
should be 15 labels. In the future, we might consider it to scale the 
proposed solution to evolve multi-label learning algorithm. For instance, we must impose a constraint in the grammar that specifies the use of PCC only 
if the number of labels is less than 15. We can also specify a time budget for all MLC algorithms, depending on the size of the dataset. 
This will consequently limit the effectiveness of the PCC algorithm in more complex types of data.
\end{enumerate}


\subsection{Monte-Carlo Classifier Chains}\label{mlc:mcc}

The algorithms based on Monte-Carlo classifier chains, MCC and M2CC \citep{Read2013,Read2014}, apply classifier chains with Monte Carlo optimization, using a maximum number of inference and chain-order trials. 
MCC has a tractable label prediction scheme  only at the test time (MCC), whereas M2CC performs an additional search for the optimal
chain sequence at the training time. 
\textbf{Parameters:}
\begin{itemize}
 \item \underline{Base classifier (\emph{bc})[-W] }: It can be any classifier from WEKA.
  \item \underline{Inference Iterations (\emph{ii})[-Iy]}: The number of iterations to search the output space at test time. This parameter is bounded by the values in the interval:
 $\{ii \in \mathbb{Z} \text{ } | \text{ } 1 < ii \leq 100 \}$. 
  \\ \textbf{Default value: }10.
 
  \item \underline{Chain Iterations (\emph{chi})[-Is]}:  The number of iterations to search the chain space at training time. This parameter is bounded by the values in the interval:
 $\{chi \in \mathbb{Z} \text{ } | \text{ } 1 < chi \leq 1500 \}$. It can also take the value zero and the MCC algorithm is used instead of M2CC. This will happen with 50\% of probability, i.e.,
 MCC and M2CC have the same chances of being selected.
  \\ \textbf{Default value: }0.
 
 \item \underline{Payoff function (\emph{pof})[-P])}: It sets the payoff function to evaluate the chains when performing the search. 
 It can take 23 values: 
 1. Accuracy; 2. Jaccard index; 3. Hamming score; 4. Exact match; 5. Jaccard distance; 6. Hamming loss; 7. Zero One loss; 8. Harmonic score; 9. One error; 
 10. Rank loss; 11. Average precisio; 12. Log Loss limited by the number of labels; 13. Log loss limited by the number of instances;
 14. Micro Precision;  15. Micro Recall; 16. Macro Precision; 17. Macro Recall; 18. $F_1$ micro averaged; 19. $F_1$ macro averaged by example; 20.
 $F1$ macro averaged by label; 21. AUPRC macro averaged; 22. AUROC macro averaged; 23. Levenshtein distance.
  \\ \textbf{Default value: }Exact match.

  
\end{itemize}


There are no dependencies/constraints between the parameters in MCC and M2CC. Additionally, we studied the range of the parameters in the works of Read \emph{et al.}~\citep{Read2013,Read2014}. However, the authors did not employ a proper parameter tuning
at the single-label level, neither at the multi-label level. In the work of \cite{Read2014}, a search is performed
to find the proper number of chain iterations in accordance to the payoff function. We are using part of this study to define the range of the parameters.


\subsection{Population of Monte-Carlo Classifier Chains}\label{mlc:pmcc}

The population of Monte-Carlo classifier chains (PMCC) \citep{Read2013,Read2014} is an algorithm that has similar properties when compared to MCC and M2CC. However, it is considered an extension of both algorithms.
The difference is that PMCC creates a population of $M$ chains at training time (from $Is$ candidate chains, using Monte Carlo sampling), and uses all of them at test time.
This is not a typical majority-vote ensemble algorithm.  
The simulated annealing search \citep{Kirkpatrick1983} can also be applied to the chain structures (produced by MCC or M2CC) in order to find the best one. 

\textbf{Parameters:}
\begin{itemize}
 \item \underline{Base classifier (\emph{bc})[-W] }: It can be any classifier from WEKA.
 
   \item \underline{Inference Iterations (\emph{ii})[-Iy]}: The number of iterations to search the output space at test time. This parameter is bounded by the values in the interval:
 $\{ii \in \mathbb{Z} \text{ } | \text{ } 1 < ii \leq 100 \}$. 
  \\ \textbf{Default value: }10.
 
  \item \underline{Chain Iterations (\emph{chi})[-Is]}:  The number of iterations to search the chain space at training time. This parameter is bounded by the values in the interval:
 $\{chi \in \mathbb{Z} \text{ } | \text{ } 50 < chi \leq 1500 \}$.  
 \\ \textbf{Default value: }50.
 
 \item \underline{Beta ($\beta$)[-B]}: It sets the factor with which the temperature 
 (and thus the acceptance probability of steps in the wrong direction in the search space) is decreased in each iteration of the simulated annealing search. This parameter
 is bounded by the interval: $\{\beta \in \mathbb{Z} \text{ } | \text{ } 0.01 \leq \beta \leq 0.99 \}$.
  \\ \textbf{Default value: }0.03.
 
 \item \underline{Temperature switch \emph{(ts)}[-O]}: It sets the use of simulated annealing search and, when it is activated, it cools
 the chain down over time (from the beginning of the chain). It may take the values zero (0) or one (1).
 The value zero (0) means that no temperature is used, i.e., the parameter $\beta$ is ignored internally by PMCC.  If using $ts=1$, this sets the use of the $\beta$ constant. 
  \\ \textbf{Default value: }0.
 
  \item \underline{Population size (\emph{ps})[-M]}: It sets the population size.  It should be always smaller than the total number of chains evaluated (Is).
  This parameter takes one of the values defined by the following interval: \\ $\{ps \in \mathbb{Z} \text{ } | \text{ } 1 \leq ps \leq 50 \}$.
   \\ \textbf{Default value: }10.
  
   \item \underline{Payoff function (\emph{pof})[-P])}: It sets the payoff function to evaluate the chains when performing the search. 
 It can take 23 values: 
 1. Accuracy; 2. Jaccard index; 3. Hamming score; 4. Exact match; 5. Jaccard distance; 6. Hamming loss; 7. Zero One loss; 8. Harmonic score; 9. One error; 
 10. Rank loss; 11. Average precisio; 12. Log Loss limited by the number of labels; 13. Log loss limited by the number of instances;
 14. Micro Precision;  15. Micro Recall; 16. Macro Precision; 17. Macro Recall; 18. $F_1$ micro averaged; 19. $F_1$ macro averaged by example; 20.
 $F1$ macro averaged by label; 21. AUPRC macro averaged; 22. AUROC macro averaged; 23. Levenshtein distance.
  \\ \textbf{Default value: }Exact match.

\end{itemize}

 \textbf{Dependencies/Constraints:}
 \begin{enumerate}
 \item The parameter ``population size'' must be smaller than the parameter ``chain iterations''.
 \end{enumerate}

Again, we studied the range of the parameters in the works of \cite{Read2014}. However, the authors did not employ a proper parameter tuning
at the single-label level, neither at the multi-label level. During the work of \cite{Read2014}, a search is performed
to find the proper number of chain iterations in accordance to the payoff function. We are using part of this study to define the range of the parameters.
Nevertheless, parameters $\beta$, temperature and population size are not properly studied for the multi-label scenario. 


\subsection{Classifier Trellis}\label{mlc:ct}

The classifier trellis (CT) algorithm \citep{Read2015} builds classifier chains in a trellis structure (rather than a cascaded chain). 
It is possible to set the width and type/connectivity of the trellis, and optionally to change the payoff function which guides the placement of nodes (labels) within the trellis.
\textbf{Parameters:}
\begin{itemize}
 \item \underline{Base classifier (\emph{bc})[-W] }: It can be any classifier from WEKA.
 
  \item \underline{Width (\emph{w})[-H]}: it determines the width of the trellis (0 for chain, i.e., $w = L$; 
  -1 for a square trellis, i.e., $w = \sqrt{L}$, always using the default floor function to convert it to an integer value). 
  Thus, the trellis structure will always have $w$ rows and $L$ nodes, in total, connected using directed edges.
   \\ \textbf{Default value: }-1.
  
 \item \underline{Dependency type (\emph{dp})[-X]}: The way to measure and find the label dependencies. It may take nine categorical values:
  1. C (co-occurrence counts);   2. I (mutual information); 3. Ib (mutual information using binary approximation);
  4. Ibf (Mutual information using fast binary approximation); 5. H (Conditional information);
  6. Hbf ( Conditional information using fast binary approximation); 7. X (Chi-squared);
  8. F (Frequencies); 9. No label dependence.
   \\ \textbf{Default value: }Ibf.
  
   \item \underline{Inference Iterations (\emph{ii})[-Iy]}: The number of iterations to search the output space at test time. This parameter is bounded by the values in the interval:
 $\{ii \in \mathbb{Z} \text{ } | \text{ } 1 \leq ii \leq 100 \}$. 
  \\ \textbf{Default value: }10.
 
  \item \underline{Chain Iterations (\emph{chi})[-Is]}:  The number of iterations to search the chain space at train time. This parameter is bounded by the values in the interval:
 $\{chi \in \mathbb{Z} \text{ } | \text{ } 1 < chi \leq 1500 \}$. 
  \\ \textbf{Default value: }0.
 
  \item \underline{Density (\emph{d})[-L]}: It determines the neighborhood density (the number of neighbors for each node in the trellis). 
 The default value for the density parameter
 is one (1), and zero (0) indicates a BR classifier. Thus, this parameter is not allowed to take the value zero, 
 being restricted by the interval:  $\{d \in \mathbb{Z} \text{ } | \text{ } 1 \leq d \leq \sqrt{L} + 1 \}$, where $L$ is the total number of labels.
  \\ \textbf{Default value: }1.
 
  \item \underline{Payoff function (\emph{pof})[-P])}: It sets the payoff function to evaluate the chains when performing the search. 
 It can take 23 values: 
 1. Accuracy; 2. Jaccard index; 3. Hamming score; 4. Exact match; 5. Jaccard distance; 6. Hamming loss; 7. Zero One loss; 8. Harmonic score; 9. One error; 
 10. Rank loss; 11. Average precisio; 12. Log Loss limited by the number of labels; 13. Log loss limited by the number of instances;
 14. Micro Precision;  15. Micro Recall; 16. Macro Precision; 17. Macro Recall; 18. $F_1$ micro averaged; 19. $F_1$ macro averaged by example; 20.
 $F1$ macro averaged by label; 21. AUPRC macro averaged; 22. AUROC macro averaged; 23. Levenshtein distance.
  \\ \textbf{Default value: }Exact match.
  
 \end{itemize}
 
 \textbf{Dependencies/Constraints:}
 \begin{enumerate}
  \item If the width $w = L$ ($w=0$), the density $d = 1$. Otherwise, if $w = \sqrt{L}$ (w = -1), the density $d$ should be $\sqrt{L} + 1$, at most, i.e., $d \leq \sqrt{L} + 1 $.
 \end{enumerate}


\subsection{Conditional Dependency Networks}\label{mlc:cdn}

The conditional dependency networks (CDN) algorithm \citep{Guo2011} builds a fully connected undirected network, where each node (label) is connected to each other node (label). Each node is a binary classifier that predicts $p(y_j|x,y_1,...,y_{j-1},...,y_L)$. 
Then, inference is done using the Gibbs Sampling method over $I$ iterations. Additionally, the final $I_c$ iterations are used to collected 
the marginal probabilities, which become the prediction ($y[]$).

\textbf{Parameters:}
\begin{itemize}
 \item \underline{Base classifier (\emph{bc})[-W] }: It can be any classifier from WEKA.
 
  \item \underline{Iterations (\emph{i})[-I]}: The total number of iterations to perform in CDT. This parameter is restricted by the interval: $\{i \in \mathbb{Z} \text{ } | \text{ } 100 < i \leq 1000 \}$.
   \\ \textbf{Default value: }1000.
 
 \item \underline{Collection iterations (\emph{ci})[-Ic]} The number of collection iterations used to compute the output class probabilities in the Gibbs Sampling method. 
 The parameter \emph{ci} is restricted by the interval:  $\{ci \in \mathbb{Z} \text{ } | \text{ } 1 \leq ci \leq 100 \}$.
  \\ \textbf{Default value: }100.
\end{itemize}

 \textbf{Dependencies/Constraints:}
 \begin{enumerate}
  \item  The collections will happen just after $(i - ci)$ iterations. So, $i$ should be substantially greater than $ci$ in order to make the algorithm works properly.
 \end{enumerate}


\subsection{Conditional Dependency Trellis}\label{mlc:cdt}

The conditional dependency trellis (CDT) algorithm \citep{Read2015,Guo2011} is similar to the CDN approach. However, it constructs a trellis structure (like CT) instead of a fully connected network.
\textbf{Parameters:}
\begin{itemize}
 \item \underline{Base classifier (\emph{bc})[-W] }: It can be any classifier from WEKA.
 
  \item \underline{Width (\emph{w})[-H]}: it determines the width of the trellis (0 for chain, i.e., $w = L$; 
  -1 for a square trellis, i.e., $w = \sqrt{L}$, always using the default floor function to convert it to an integer value). 
  Thus, the trellis structure will always have $w$ rows and $L$ nodes, in total, connected using directed edges.
   \\ \textbf{Default value: }-1.
  
 \item \underline{Dependency type (\emph{dp})[-X]}: The way to measure and find the label dependencies. It may take nine categorical values:
  1. C (co-occurrence counts);   2. I (mutual information); 3. Ib (mutual information using binary approximation);
  4. Ibf (Mutual information using fast binary approximation); 5. H (Conditional information);
  6. Hbf ( Conditional information using fast binary approximation); 7. X (Chi-squared);
  8. F (Frequencies); 9. None (Using empty).
   \\ \textbf{Default value: }None.
  
  \item \underline{Density (\emph{d})[-L]}: It determines the neighborhood density (the number of neighbors for each node in the trellis). 
 The default value for the density parameter
 is one (1), and zero (0) indicates a BR classifier. Thus, this parameter is not allowed to take the value zero, 
 being restricted by the interval:  $\{d \in \mathbb{Z} \text{ } | \text{ } 1 \leq d \leq \sqrt{L} + 1 \}$, where $L$ is the total number of labels.
  \\ \textbf{Default value: }1.
  
  \item \underline{Iterations (\emph{i})[-I]}: The total number of iterations to perform in CDT. This parameter is restricted by the interval: $\{i \in \mathbb{Z} \text{ } | \text{ } 100 < i \leq 1000 \}$.
   \\ \textbf{Default value: }1000.
 
 \item \underline{Collection iterations (\emph{ci})[-Ic]} The number of collection iterations used to compute the output class probabilities in the Gibbs Sampling method. 
 The parameter \emph{ci} is restricted by the interval:  $\{ci \in \mathbb{Z} \text{ } | \text{ } 1 \leq ci \leq 100 \}$.
  \\ \textbf{Default value: }100.
 
\end{itemize}

 \textbf{Dependencies/Constraints:}
 \begin{enumerate}
   \item If the width $w = L$ ($w=0$), the density $d = 1$. Otherwise, if $w = \sqrt{L}$ (w = -1), the density $d$ should be $\sqrt{L} + 1$, at most, i.e., $d \leq \sqrt{L} + 1 $.
  \item  The collections will happen just after $(i - ci)$ iterations. So, $i$ should be substantially greater than $ci$ in order to make the algorithm works properly.
 \end{enumerate}


\subsection{Four-Class Pairwise Classification}\label{mlc:fw}

The Four-class PairWise Classification (FW) algorithm \citep{Read2016} trains a multi-class base classifier for each pair of labels. Thus, the number of classifiers is $\frac{(L*(L-1))}{2}$ in total (where L is the number of labels),
each one with four possible class values (00,01,10,11) representing the possible combinations of relevant~(1)/irrelevant~(0) values for each label in the label pair. 
It uses a voting and a threshold scheme at testing time where, e.g., 01 from pair $jk$ gives one vote to label $k$ and
any label with a number of votes above the threshold is considered relevant. It uses the same threshold specified in the Section \ref{searchspacestudy} to define 
the relevance of a label.
\textbf{Parameters:}
\begin{itemize}
 \item \underline{Base classifier (\emph{bc})[-W] }: It can be any classifier from WEKA.
\end{itemize}

 There are no dependencies/constraints in FW.
 

\subsection{Ranking and Threshold}\label{mlc:rt}

The ranking and threshold (RT) algorithm \citep{Read2010} duplicates each multi-labeled example, and assigns one of the labels (only) to each copy. After that, it trains a regular multi-class base classifier. At test time, a threshold separates relevant from irrelevant labels using the posterior probability for each class value (i.e., label). 
It uses the same threshold specified in the Section \ref{searchspacestudy} to define the relevance of a label.
\textbf{Parameters:}
\begin{itemize}
 \item \underline{Base classifier (\emph{bc})[-W] }: It can be any classifier from WEKA.
\end{itemize}

 There are no dependencies/constraints in RT.


\subsection{Label Combination}\label{mlc:lc}

The label combination (LC) algorithm \citep{Tsoumakas2010}, also knows as label powerset (LP),  treats each label combination as a single class in a multi-class learning scheme. 
The set of possible values of each class is the powerset of the set of labels.

\begin{itemize}
 \item \underline{Base classifier (\emph{bc})[-W] }: It can be any classifier from WEKA.
\end{itemize}

 There are no dependencies/constraints in LC.


\subsection{Pruned Sets}\label{mlc:ps}

The pruned sets (PS) algorithm \citep{Read2010,Read2008} was created to use the power of LC's labelset-based paradigm, without the disadvantages of such algorithm. In order to to this, this algorithm has two important steps: a pruning step and a
label-set subsampling step. The pruning step removes infrequently occurring
label sets from the training data. This removes unnecessary complexity from the LC-transformed data by reducing the number of labelsets. 
Nevertheless, PS does not simply discard the pruned examples.
Instead of doing that, PS subsamples the labelsets of these examples for label subsets which occur more frequently in the
training data. It then attaches these label sets to the example, creating new examples and reintroducing them into the training.
It subsamples these labelsets $pv$ times to produce $pv$ new examples, where $pv$ is the pruning value (defined in the followed items).



After these steps, it trains a standard LC classifier. The idea of the algorithm is to reduce the number of unique class values that 
would otherwise need to be learned by LC. PS achieves its best performance when used in an Ensemble (e.g., EnsembleML). \textbf{Parameters:}
\begin{itemize}
 \item \underline{Base classifier (\emph{bc})[-W] }: It can be any classifier from WEKA.
 
 \item \underline{Pruning value (\emph{pv})[-P]}: It defines an infrequent labelset as one which occurs less than $p$ times in the data. 
 $p = 0$ would mean that LC classifier is learned. Thus, this parameter is bounded by the  following interval: $\{pv \in \mathbb{Z} \text{ } | \text{ } 1 \leq pv \leq 5 \}$.
  \\ \textbf{Default value: }0.
 
  \item \underline{Subsampling value (\emph{sv})[-N]}: 
  The label set of each pruned example (in accordance to the examples pruned by the use of the previous parameter, i.e., the pruning value) becomes a candidate for label-set subsampling.
  The PS algorithm subsamples the label sets of pruned examples to create examples which do meet the pruning criterion.
  So, the subsample value defines the (maximum) number of frequent labelsets to subsample from the infrequent labelsets. This parameter is bounded by the 
 following interval: $\{sv \in \mathbb{Z} \text{ } | \text{ } 0 \leq sv \leq 5 \}$.
  \\ \textbf{Default value: }0.
 
 \end{itemize}
 
 There are no dependencies/constraints between the parameters of PS. 
 Additionally, there is a proper study in the work of \cite{Read2010} about the range of these two parameters.
 

\subsection{Pruned Sets with Threshold}\label{mlc:pst}

The pruned sets algorithm with a threshold (PSt) \citep{Read2010,Read2008,Read2008b}, which is a modification of PS that can form new label sets at classification (i.e., test) time by using a threshold function.
Given the posterior of the label classes (combinations) and the number of labels, it returns the distribution across labels.
Using the threshold (defined in the Section \ref{searchspacestudy}) could make the algorithm to predict labelsets not seen in the training set, differently from PS. 
\textbf{Parameters:}
\begin{itemize}
 \item \underline{Base classifier (\emph{bc})[-W] }: It can be any classifier from WEKA.
 
 \item \underline{Pruning value (\emph{pv})[-P]}: It defines an infrequent labelset as one which occurs less than $p$ times in the data. 
 $p = 0$ would mean that LC classifier is learned. Thus, this parameter is bounded by the  following interval: $\{pv \in \mathbb{Z} \text{ } | \text{ } 1 \leq pv \leq 5 \}$.
 \\ \textbf{Default value: }0.
 
  \item \underline{Subsampling value (\emph{sv})[-N]}: 
  The label set of each pruned example (in accordance to the examples pruned by the use of the previous parameter, i.e., the pruning value) becomes a candidate for label-set subsampling.
  The PSt algorithm subsamples the label sets of pruned examples to create examples which do meet the pruning criterion.
  So, the subsample value defines the (maximum) number of frequent labelsets to subsample from the infrequent labelsets. This parameter is bounded by the 
 following interval: $\{sv \in \mathbb{Z} \text{ } | \text{ } 0 \leq sv \leq 5 \}$.
 \\ \textbf{Default value: }0.
 
 \end{itemize}

There are no dependencies/constraints between the parameters of PSt. Additionally,
there is a proper study in the work of \cite{Read2010} about the range of these two parameters ($pv$ and $sv$). The parameters are the same of PS. The main thing that is
changed in PSt when compared to PS occurs at the test time.


\subsection{Random k-Label Pruned Sets}\label{mlc:rakel}

The RAndom k-labEL Pruned Sets (RAkEL) algorithm \citep{Tsoumakas2011,Read2010} randomly draws $M$ subsets of labels, each with $k$ labels, from the set of labels, and trains PS upon each one. 
Finally, it combines label votes from the PS classifiers to get a label-vector prediction.
\textbf{Parameters:}
\begin{itemize}
 \item \underline{Base classifier (\emph{bc})[-W] }: It can be any classifier from WEKA.

 \item \underline{Pruning value (\emph{pv})[-P]}: It prunes an infrequent labelset when it occurs less than $pv$ times in the data. 
 $pv = 0$ means that LC classifier is learned. Thus, this value is not allowed for RAkEL, which makes this parameter being bounded by the  following interval: \\ $\{pv \in \mathbb{Z} \text{ } | \text{ } 1 \leq pv \leq 5 \}$.
 \\ \textbf{Default value: }0.

  \item \underline{Subsampling value (\emph{sv})[-N]}: 
  The label set of each pruned example (in accordance to the examples pruned by the use of the previous parameter, i.e., the pruning value) becomes a candidate for label-set subsampling.
  This version of RAKEL in MEKA subsamples the label sets of pruned examples to create examples which do meet the pruning criterion.
  So, the subsample value defines the (maximum) number of frequent labelsets to subsample from the infrequent labelsets. This parameter is bounded by the 
 following interval: $\{sv \in \mathbb{Z} \text{ } | \text{ } 0 \leq sv \leq 5 \}$.
 \\ \textbf{Default value: }0.
 
  \item \underline{Number of labels for each subset (\emph{les})[-k]}:  It defines the number of labels in each label subset.
 This parameter should be bounded by the interval~\citep{Madjarov2012}: \\ $\{les \in \mathbb{Z} \text{ } | \text{ } 1 \leq les \leq \frac{L}{2}$~\}, where $L$ is the number of labels.
 \\ \textbf{Default value: }3.
 
   \item \underline{Number of subsets to run in an ensemble} (\emph{sre})[-M]): This parameter controls the number of models to build in a ensemble and
  take values in accordance to the following interval~\citep{Madjarov2012}: \\ $\{sre \in \mathbb{Z} \text{ } | \text{ } 2 \leq sre \leq min(2 \cdot L ,100)$\}, where L is the number of labels.
  \\ \textbf{Default value: }10.


\end{itemize}

There are no dependencies/constraints between the parameters of RAkEL.  
Additionally,
we followed the work of \cite{Read2010} about the range of the subsampling and pruning values. The other two parameters (number of labels in each subset and 
 number of models to build in a ensemble) were defined in accordance 
to the work of \cite{Madjarov2012}.


\subsection{Random k-Label Disjoint Pruned Sets}\label{mlc:rakeld}

The RAndom k-labEL Disjoint Pruned Sets (RAkELd) algorithm \citep{Tsoumakas2011,Read2010} takes a random partition of labels, but unlike RAkEL the labelsets are disjoint/non-overlapping subsets. \textbf{Parameters:}
\begin{itemize}
 \item \underline{Base classifier (\emph{bc})[-W] }: It can be any classifier from WEKA.

 \item \underline{Pruning value (\emph{pv})[-P]}: It prunes an infrequent labelset when it occurs less than $p$ times in the data. 
 $pv = 0$ means that LC classifier is learned. Thus, this value is not allowed for RAkEL, which makes this parameter being bounded by the  following interval: \\ $\{pv \in \mathbb{Z} \text{ } | \text{ } 1 \leq pv \leq 5 \}$.
 \\ \textbf{Default value: }0.
 
  \item \underline{Subsampling value (\emph{sv})[-N]}: 
  The label set of each pruned example (in accordance to the examples pruned by the use of the previous parameter, i.e., the pruning value) becomes a candidate for label-set subsampling.
  The version of RAKEd in MEKA subsamples the label sets of pruned examples to create examples which do meet the pruning criterion.
  So, the subsample value defines the (maximum) number of frequent labelsets to subsample from the infrequent labelsets. This parameter is bounded by the 
 following interval: $\{sv \in \mathbb{Z} \text{ } | \text{ } 0 \leq sv \leq 5 \}$.
 \\ \textbf{Default value: }0.

%
   \item \underline{Number of subsets to run in an ensemble} (\emph{sre})[-M]): This parameter controls the number of models to build in a ensemble and
  take values in accordance to the following interval~\citep{Madjarov2012}: \\ $\{sre \in \mathbb{Z} \text{ } | \text{ } 2 \leq sre \leq min(2 \cdot L ,100)$\}, where L is the number of labels.
   \\ \textbf{Default value: }10.

  \end{itemize}

There are no dependencies/constraints between the parameters of RAkELd.  Additionally,
we followed the work of \cite{Read2010} again to set the range of the subsampling and pruning values. The other two parameters (number of labels in each subset and 
 number of models to build in a ensemble) were defined in accordance 
to the work of \cite{Madjarov2012}.

\subsection{Multi-Label Back Propagation Neural Network} \label{mlc:bpnn}

The multi-label back propagation neural network (ML-BPNN) algorithm \citep{Zhang2006,Read2014deep} is a standard Back-Propagation Neural Network \citep{Rumelhart1986} with multiple outputs that correspond to multiple labels.
That is, each node in the output layer corresponds to a different class label.
\textbf{Parameters:}
\begin{itemize}
 \item \underline{Number of epochs (\emph{ne})[-E]}: It is the number of iterations to train the neural network. 
 It is restricted by the interval: $\{ne \in \mathbb{Z} \text{ } | \text{ } 10 \leq ne \leq 1000 \}$.
  \\ \textbf{Default value: }100.
 
 \item \underline{Number of hidden units (\emph{nhu})[-H]}: It defines the number of hidden units in the neural network. It is import to mention that
 the version of ML-BPNN in MEKA is limited to one hidden layer with $nhu$ hidden units. This parameter
  takes values in proportion to the number of attributes (received as input). Thus, the number of hidden units of the network
 can vary from 20\% to 100\% of the number of attributes: $\{nhu \in \mathbb{Z} \text{ } | \text{ } 0.2 \cdot number\_of\_attributes \leq nhu \leq number\_of\_attributes \}$.
 The proportion will always be rounded to the nearest integer.
  \\ \textbf{Default value: }10.

  \item \underline{Learning rate (\emph{lrt})[-r]}: The amount by which the weights are updated during training. 
  It is restricted by the interval: \\ $\{lr \in \mathbb{R} \text{ } | \text{ } 0.001 \leq lr \leq 0.1 \}$.
   \\ \textbf{Default value: }0.1.
  \item \underline{Momentum (\emph{m})[-m]}: It is applied to the weights during updating. 
  It is restricted by the interval: $\{m \in \mathbb{R} \text{ } | \text{ } 0.1 \leq m \leq 0.8 \}$.
   \\ \textbf{Default value: }0.1.

 \end{itemize}

There are no dependencies/constraints between the parameters of ML-BPNN. Additionally,
the range of values for the parameters number of epochs, momentum and learning rate were set following the work of \cite{Read2014deep}.
The only parameter which was defined based on a different work~\citep{Zhang2006} was the number of hidden units, $nhu$.

\section{Search Space -- MLC Meta-Algorithms}
\label{meta-mlc}

In this section, we describe the \emph{search space} of multi-label meta-algorithms in MEKA. 
It is important to say that some of the multi-label classifiers (presented in the last section) do not perform very well when used as the multi-label base classifier
in a meta classifier.
This is due to the poor scalability of such combination (meta multi-label and base multi-label). Examples of algorithms that would not scale up well are:
MCC, PCC, PMCC, CDN and CDT (these last two algorithms involve Gibbs sampling, which may be too expensive in an ensemble),  RAkEL and RAkELd  (these two algorithms are ensembles by themselves, and using an ensemble as base classifier would lead to a very slow ensemble of ensembles).
This must be considered in the grammar or directly in the execution of the algorithm (i.e, setting a time budget for such algorithms when they are used at the multi-label base level).


%


\subsection{Subset Mapper}\label{mlc:sm}

The subset mapper (SM) algorithm \citep{Schapire1999} maps the output of a multi-label classifier to a known label combination using the Hamming distance, i.e., it checks what label combination (label subsets) from the training set has the closest distance to the predicted label combination on the 
test instance using probability distribution of the label subset for this instance. 
In order to do that, SM transforms the probability distribution array of the label subset in a binary array. 
For each label subset in the training set (also represented by a binary array), it calculates the Hamming distance to the binary probability distribution array, outputting the closest label subset to the predicted distribution array.
SM will map this label subset to this particular test instance.
\textbf{Parameters:}
\begin{itemize}
 \item \underline{Multi-label classifier(\emph{mlc})[-W]}: The multi-label algorithm that creates a model at the multi-label classification level.
\end{itemize}

\textbf{Dependencies/Constraints:}
\begin{enumerate}
 \item The multi-label classification algorithm can be any one described in the Section \ref{mlc}.
\end{enumerate}

\subsection{Bagging of Multi-Label Classifiers}\label{mlc:bagging}

The bagging of multi-label classifiers (BaggingML) \citep{Read2010} is an algorithm that combines several multi-label classifiers using Bootstrap AGGregatING (Bagging) \citep{Breiman1996}.
It randomly sets weights higher than zero to certain instances, on only those instances are chosen for the bag.
The parameter ``bag percent size'' is then not used as the number of instances in the bag is just based on the weight values.
Thus, the members of the ensemble could have 100\% of the instances if all of them have a weight assigned. 
\textbf{Parameters:}
\begin{itemize}
 \item \underline{Multi-label classifier(\emph{mlc})[-W]}: The multi-label algorithm that creates a model at the base multi-label classification level.
 \item \underline{Number of  iterations (\emph{i})[-I]}: The number of iterations to perform, i.e., the number of members in the ensemble. This parameter is restricted by the interval: $\{i \in \mathbb{Z} \text{ } | \text{ } 10 \leq i \leq 50 \}$.
The range of the number of iterations for this ensemble algorithm was defined by Read's thesis \citep{Read2010}. 
\\ \textbf{Default value: }10.
 \end{itemize}

\textbf{Dependencies/Constraints:}
\begin{enumerate}
 \item The multi-label classification algorithm can be any one described in the Section \ref{mlc}, except for BCC, which is not suitable for this meta-learner.
\end{enumerate}



\subsection{Bagging of Multi-Label Classifiers with Duplicates}\label{mlc:baggingdup}

BaggingML with duplicates (BaggingMLDup) \citep{Read2010} is an algorithm that also combines several multi-label classifiers using Bootstrap AGGregatING. 
However, it uses the parameter ``bag size percent'' to define a specific number of instances for each member (classifier) of the ensemble.
After that, it randomly samples instances, being able to sample the same instance ($duplicates$) for the bag.
This algorithm does not use any weight to select the instances for the members of the ensemble.
\textbf{Parameters:}
\begin{itemize}
 \item \underline{Multi-label classifier (\emph{mlc})[-W]}: The multi-label algorithm that creates a model at the multi-label classification level.
 \item \underline{Bag size percent (\emph{bsp})[-P]}: The size of the bag in percentage of the training set size (number of training instances)  and it is defined by the interval: $\{bsp \in \mathbb{Z} \text{ } | \text{ } 10 \leq bsp \leq 100 \}$.
 \\ \textbf{Default value: }67.
 \item \underline{Number of  iterations (\emph{i})[-I]}: The number of iterations to perform, i.e., the number of members in the ensemble. This parameter is restricted by the interval: $\{i \in \mathbb{Z} \text{ } | \text{ } 10 \leq i \leq 50 \}$.
 \\ \textbf{Default value: }10.

\end{itemize}

\textbf{Dependencies/Constraints:}
\begin{enumerate}
 \item The multi-label classification algorithm can be any one described in the Section \ref{mlc}, except for BCC, which is not suitable for this meta-learner.
\end{enumerate}

The range of the number of iterations for this ensemble algorithm was defined by Read's thesis \citep{Read2010}.
The parameter ``bag size percent'' is defined in the  MEKA documentation.


\subsection{Ensemble of Multi-Label Classifiers}\label{mlc:ensemble}

The ensemble of multi-label classifiers (EnsembleML) \citep{Read2010} is an algorithm that combines several multi-label classifiers in a simple-subset ensemble.
This algorithm is very similar to BaggingMLDup. The only difference is that BaggingMLDup allows 
sampling with replacement for each model, whereas EnsembleML  uses sampling without replacement. 

\textbf{Parameters:}
\begin{itemize}
 \item \underline{Multi-label classifier (\emph{mlc})[-W]}: The multi-label algorithm that creates a model at the base multi-label classification level.
 \item \underline{Bag size percent (\emph{bsp})[-P]}: The size of the bag in percentage of the training size (number of training instances) and it is defined by the interval: $\{bsp \in \mathbb{Z} \text{ } | \text{ } 52 \leq bsp \leq 72 \}$.
  \\ \textbf{Default value: }67.
 \item \underline{Number of  iterations (\emph{i})[-I]}: The number of iterations to perform, i.e., the number of members in the ensemble. This parameter is restricted by the interval: $\{i \in \mathbb{Z} \text{ } | \text{ } 10 \leq i \leq 50 \}$.
 \\ \textbf{Default value: }10.

\end{itemize}

\textbf{Dependencies/Constraints:}
\begin{enumerate}
 \item The multi-label classification algorithm can be any one described in the Section \ref{mlc}, except for BCC, which is not suitable for this meta-learner.
\end{enumerate}

The range of the number of iterations for this ensemble algorithm was defined by Read's thesis \citep{Read2010}.
Additionally, in his thesis, the author mentioned that they found that values around 62\% are the best ones for the parameter ``bag size percent'' in a ensemble without replacement, which is the case. 
Thus, we are trying to set the range for this parameter introducing lower and upper bounds close to this value (10\% smaller and 10\% greater).


\subsection{Random Subspace Multi-Label}\label{mlc:rsml}

The random subspace multi-label (RSML) algorithm \citep{Breiman2001} combines several multi-label classifiers in an ensemble where the attribute space and the instance space used for building each model are random subsets from the original space. 
In other words, RSML subsamples the attribute space and instance space randomly for each ensemble member. 
Basically, it is a generalized version of Random Forests. Additionally, it is computationally cheaper than EnsembleML for the same number of models
in the ensemble and the same value of bag size percent.

\textbf{Parameters:}
\begin{itemize}
 \item \underline{Multi-label classifier (\emph{mlc})[-W]}: The multi-label algorithm that creates a model at the base multi-label classification level.
 \item \underline{Bag size percent (\emph{bsp})[-P]}: The size of the bag in percentage of the training set size (number of training instances), and it is defined by the interval: $\{bsp \in \mathbb{Z} \text{ } | \text{ } 10 \leq bsp \leq 100 \}$.
 \\ \textbf{Default value: }67.
 \item \underline{Number of  iterations (\emph{i})[-I]}: The number of iterations to perform, i.e., the number of members in the ensemble. This parameter is restricted by the interval: $\{i \in \mathbb{Z} \text{ } | \text{ } 10 \leq i \leq 50 \}$.
 \\ \textbf{Default value: }10.
 \item \underline{Attribute percent (\emph{ap})[-A]}: The size of the attribute space, as a percentage of total attribute space size (number of attributes). This parameter is bounded by the following interval:
 $\{ap \in \mathbb{Z} \text{ } | \text{ } 10 \leq ap \leq 100 \}$.
 \\ \textbf{Default value: }50.
\end{itemize}

\textbf{Dependencies/Constraints:}
\begin{enumerate}
 \item The multi-label classification algorithm can be any one described in the Section \ref{mlc}.
\end{enumerate}

The range of the number of iterations for this ensemble algorithm was defined based on Read's thesis \citep{Read2010}.
The range of values for this parameter also considers scalability issues as we need to run a multi-label algorithm many times in an ensemble.
The parameter ``bag size percent'' is defined in the MEKA documentation.
The attribute percentage was set in accordance to the single-label version for the same algorithm. This was done because there is not any work that studies this algorithm
for multi-label classification.


\subsection{Expectation Maximization}\label{mlc:em}

In the expectation maximization (EM) algorithm \citep{Dempster1977}, a specified multi-label classifier is built on the training data. 
This model is then used to classify the training data. The confidence with which instances are classified is used to reweight them. 
This data is then used to retrain the classifier. This cycle continues (`EM'-style) for $I$ iterations. The final model is used to 
classify the test data. Because of the weighting, it is advised to use a classifier which gives good confidence (probabilistic) outputs.  \textbf{Parameters:}
\begin{itemize}
 \item \underline{Multi-label classifier (\emph{mlc})[-W]}:  The multi-label algorithm that creates a model at the multi-label classification level.
 \item \underline{Number of  iterations (\emph{i})[-I]}: The number of iterations to perform. This parameter is restricted by the interval: $\{i \in \mathbb{Z} \text{ } | \text{ } 10 \leq i \leq 50 \}$.
 \\ \textbf{Default value: }10.

\end{itemize}

\textbf{Dependencies/Constraints:}
\begin{enumerate}
 \item The classifier at the base multi-label classification level should be capable to produce probabilistic predictions. However, in our preliminary tests, 
 most multi-label classification algorithms described in Section \ref{mlc}  were suitable for this meta-learner, except for PMCC.
 Thus, we will use all the suitable algorithms in accordance to these experiments at the base multi-label classification level.
\end{enumerate}

The range of the number of iterations for this ensemble algorithm was defined based on the Read's thesis \citep{Read2010}.
The range of values for this parameter also considers scalability issues as we need to run a multi-label algorithm many times in an ensemble.
This was done because there is not an appropriate work that studies this algorithm
for multi-label classification.


\subsection{Classification Maximization}\label{mlc:cm}

The classification maximization (CM) algorithm \citep{Dempster1977,Read2016} trains a classifier with labeled and unlabeled data (semi-supervised) learning using the Classification Expectation algorithm, which is a hard version of EM algorithm, as it does not update the instance weights using (a product factor of) the probability distribution produced by the classifier.
Instead, it sets to zero (0.0) or one (1.0) the weight of any instance in the dataset.
Unlike EM, it can use any classifier, not necessarily one that gives good probabilistic outputs.  
\textbf{Parameters:}
\begin{itemize}
 \item \underline{Multi-label classifier (\emph{mlc})[-W]}:  The multi-label algorithm that creates a model at the base multi-label classification level.
 \item \underline{Number of  iterations (\emph{i})[-I]}: The number of iterations to perform. This parameter is restricted by the interval: $\{i \in \mathbb{Z} \text{ } | \text{ } 10 \leq i \leq 50 \}$.
 \\ \textbf{Default value: }10.

\end{itemize}

\textbf{Dependencies/Constraints:}
\begin{enumerate}
 \item The multi-label classification algorithm can be any one described in the Section \ref{mlc}, except for PMCC, which is not suitable for this meta-learner.
\end{enumerate}

The range of the number of iterations for this ensemble algorithm was defined based on Read's thesis \citep{Read2010}.
The range of values for this parameter also considers scalability issues as we need to run a multi-label algorithm many times in an ensemble. 
This was done because there is not any work that studies this algorithm
for multi-label classification.

\begin{table*}[!t]

\begin{center}
 \caption{Overview of the employed single-label classification (SLC) algorithms from the WEKA software.$^*$}
 \label{mlc:tab2}
\begingroup

\begin{tabular}{|l|l|l|l||c|c||c|c||c|c|}
\cline{5-10}
\multicolumn{4}{l}{}                                                          & \multicolumn{2}{|c|}{\textbf{Small}} & \multicolumn{2}{|c|}{\textbf{Medium}} & \multicolumn{2}{|c|}{\textbf{Large}} \\\hline
\textbf{id} & \textbf{Algorithm Name}                        & \textbf{Acronym}      & \textbf{Category}     & \textbf{Used?}          & \textbf{\#HP}          & \textbf{Used?}            & \textbf{\#HP}           & \textbf{Used?}           & \textbf{\#HP}           \\\hline                               

 1 & \hyperref[slc:jrip]{JRip}                          & JRip                           & Rules         & Y             & 4               & Y               & 4                & Y              & 4                \\ \hline  
 2 & \hyperref[slc:knn]{K-Nearest Neighbors}            & KNN                            & Lazy          & Y             & 3               & Y               & 3                & Y              & 3                \\ \hline  
 3 & \hyperref[slc:lr]{Logistic Regression}                                & LR                             & Functions     & Y             & 1               & Y               & 1                & Y              & 1                \\ \hline  
 4 & \hyperref[slc:nb]{Na\"ive Bayes}                                      & NB                             & Bayes         & Y             & 2               & Y               & 2                & Y              & 2                \\ \hline 
 5 & \hyperref[slc:rf]{Random Forest}                   & RF                             & Trees         & Y             & 3               & Y               & 3                & Y              & 3                \\ \hline  
 \hline 
 6 & \hyperref[slc:bnc]{Bayesian Network Classifier}                        & BNC                            & Bayes         & N             & -               & Y               & 1                & Y              & 1                \\ \hline  
 7 & \hyperref[slc:c4.5]{C4.5}                          & C4.5                           & Trees         & N             & -               & Y               & 8                & Y              & 8              \\ \hline 
 8 & \hyperref[slc:dt]{Decision Table}                  & DT                             & Rules         & N             & -               & Y               & 4                & Y              & 4                \\ \hline  
 9 & \hyperref[slc:k*]{K Star}                          & K*                             & Lazy          & N             & -               & Y               & 3                & Y              & 3                \\ \hline 
 10 & \hyperref[slc:lmt]{Logistic Model Trees}          & LMT                            & Trees         & N             & -               & Y               & 7                & Y              & 7                \\ \hline  
 11 & \hyperref[slc:mlp]{Multi-Layer Perceptron}        & MLP                            & Functions     & N             & -               & Y               & 6                & Y              & 6                \\ \hline  
 12 & \hyperref[slc:part]{PART}                         & PART                           & Rules         & N             & -               & Y               & 4                & Y              & 4                \\ \hline  
 13 & \hyperref[slc:reptree]{REPTree}                   & REPTree                        & Trees         & N             & -               & Y               & 3                & Y              & 3                \\ \hline 
 14 & \hyperref[slc:sgd]{Stochastic Gradient Descent}   & SGD                            & Functions     & N             & -               & Y               & 5                & Y              & 5                \\ \hline  
 15&\hyperref[slc:smo]{Sequential Minimal Optimization} & SMO                            & Functions     & N             & -               & Y               & 6              & Y              & 6                \\ \hline  
  \hline

 16 & \hyperref[slc:ds]{Decision Stump}                 & DS                             & Trees         & N             & -               & N               & -                & Y              & 0                \\ \hline 
 
 17 & \hyperref[slc:nbm]{Na\"ive  Bayes Multinomial}                        & NBM                            & Bayes         & N             & -               & N               & -                & Y              & 0                \\ \hline  
 18 & \hyperref[slc:1R]{One Rule}                       & OneR                           & Rules         & N             & -               & N               & -                & Y              & 1                \\ \hline 
 19 & \hyperref[slc:rt]{Random Tree}                                      & RT                             & Trees         & N             & -               & N               & -                & Y              & 4                \\ \hline  
 20 & \hyperref[slc:sl]{Simple Logistic}                                   & SL                             & Functions     & N             & -               & N               & -                & Y              & 3                \\ \hline  
 21 & \hyperref[slc:vt]{Voted Perceptron}               & VP                             & Functions     & N             & -               & N               & -                & Y              & 3                \\ \hline  
 22 & \hyperref[slc:0R]{Zero Rules}                     & ZeroR                          & Rules         & N             & -               & N               & -                & Y              & 0                \\ \hline 
 
 23 &  \hyperref[slc:asc]{Attribute Selection Classifier}                    & ASC                            & Preprocessing & N             & -               & N               & -                & Y              & 1   \\ \hline  
 24 &  \hyperref[slc:ada]{Ada boost M1}                                      & AdaM1                          & Meta-SLC      & N             & -               & N               & -                & Y              & 3                \\ \hline  
 25 &  \hyperref[slc:bagging]{Bagging of Single-Label Classifiers}               & Bagging                        & Meta-SLC      & N             & -               & N               & -                & Y              & 3                \\ \hline  
 26 &  \hyperref[slc:lwl]{Locally Weighted Learning}                         & LWL                            & Meta-SLC      & N             & -               & N               & -                & Y              & 2                \\ \hline  
 27 &  \hyperref[slc:rc]{Random Committee}                                  & RC                             & Meta-SLC      & N             & -               & N               & -                & Y              & 1                \\ \hline   
 28 &  \hyperref[slc:rss]{Random Subspace}                                   & RSS                            & Meta-SLC      & N             & -               & N               & -                & Y              & 3                \\ \hline  
\end{tabular}
\endgroup
\begin{tabular}{l} 
\hspace{-0.35cm} $^*$All algorithm names are clickable links to their description in this supplementary material. \hspace{4cm} \text{ }
\end{tabular}

\end{center}

\end{table*}

\section{Studying the search space of single-label classification algorithms}

In this section, we study \textbf{28} traditional (single label) classification algorithms, preprocessing algorithms and meta-algorithms from the WEKA software~\citep{Hall2009}.
This is done in order to understand the whole \emph{search space} of multi-label algorithms.
Most hyper-parameters in this section were set in accordance to the \emph{search space} definition from Auto-WEKA~\citep{Thornton2013,Kotthoff2017,Kotthoff2017b}.
The algorithms and their respective (hyper-)parameters were defined after studying the code, logs and configuration files of Auto-WEKA, which is considered a stable and robust approach for automatically selecting and configuring machine learning algorithms.

Table \ref{mlc:tab2} shows the 28 SLC algorithms used, i.e., the possible algorithms at the MLC base level and their hyper-parameters. Similar to Table~\ref{mlc:tab1}, Table~\ref{mlc:tab2} defines the SLC algorithms in terms of their names, names' acronyms and types (i.e., trees, rules, lazy, functions, Bayes, preprocessing and, meta-SLC). As before,  we show if the SLC algorithm in the row is used or not ('Y' for yes and 'N' for no) by the respective \emph{search space} in the column and how many hyper-parameters (\#HP) it has (when it is used).

\section{Search Space -- Traditional SLC Algorithms} \label{slc}

This section covers the main traditional SLC algorithms in the WEKA software \cite{Hall2009}. As already mentioned and explained, this is done as most MLC algorithms, which are described in the last two sections of this supplementary material, are from the problem transformation type, i.e., they need to employ a single-label classifier at their base level to perform the multi-label classification.


\subsection{C4.5} \label{slc:c4.5}

The algorithm for generating a C4.5 decision tree \citep{Quinlan1993}.
This algorithm can decide whether it will use the default C4.5's error-based pruning method~\citep{Bradford1998,Salzberg1994,Witten2016} or not.
If the algorithm decides to use pruning, the C4.5's pruning method is applied to the tree, and an estimation of the error rate of every subtree is done.
After that, the pruning method will replace the subtree with a leaf node if the estimated error of the leaf is lower than a threshold~\citep{Salzberg1994}. 
\textbf{Parameters:}
\begin{itemize}
  \item \underline{Confidence factor (\emph{cf})[-C]}: It is used for C4.5's error-based pruning method (smaller values incur more pruning) and is defined by the interval: \\ $\{cf \in \mathbb{R} \text{ } | \text{ } 0.0 \leq cf \leq 1.0 \}$. \\ \textbf{Default value: }0.25.
  \item \underline{Minimum number of objects (\emph{mno})[-M]}: The minimum number of instances per leaf. It can take values in the interval:  \\ $\{mno \in \mathbb{Z} \text{ } | \text{ } 1 \leq mno \leq 64 \}$. \\ \textbf{Default value: }2.
  \item \underline{Collapse tree (\emph{ct})[-O]}: It is used to decide if internal nodes will be collapsed to avoid overfitting. This parameter is used
  with C4.5's error-based pruning method to enhance the final decision tree.
  It collapses a subtree to a node only if training error of the subtree does not increase when compared to the entire tree.
  It is applied to every subtree in the tree, where subtrees are collapsed (pruned) if pruning does not increase its classification error.
  For example, if there is a subtree with two leaf nodes having the same classification on the training data, this subtree will be replaced by a single leaf.  
  It can take Boolean values (true or false).  \\ \textbf{Default value: }true.
  \item \underline{Unpruned (\emph{u})[-U]}: It decides whether pruning is performed or not. It can take Boolean values (true or false). \\ \textbf{Default value: }false.
  \item \underline{Binary splits (\emph{bs})[-B]}: It decides whether C4.5 will use binary splits on nominal attributes when building the trees.  It can take Boolean values (true or false). \\ \textbf{Default value: }false.
  \item \underline{Use MDL correction (\emph{umc})[-J]}: It decides whether the MDL correction is used when finding splits on numeric attributes.   It can take Boolean values (true or false). \\ \textbf{Default value: }true.
  \item \underline{Use Laplace (\emph{ul})[-A]}: It decides if the counts of instances at leaves are smoothed based on the Laplace correction. It can take Boolean values (true or false). \\ \textbf{Default value: }false.
  \item \underline{Subtree raising (\emph{sr})[-S]}: It is used for C4.5's error-based pruning and decides whether the algorithm will consider the subtree raising operation when pruning.  It can take Boolean values (true or false). \\ \textbf{Default value: }true.
  
 \end{itemize}
 
 \textbf{Dependencies/Constraints:}
 \begin{enumerate}
  \item If the parameter unpruned is set to ``true'', the parameters ``confidence factor'', ``collapse tree'' and ``subset raising'' are not used (omitted).
 \end{enumerate}


\subsection{Logistic Model Trees} \label{slc:lmt}

The algorithm for building logistic model trees (LMT) \citep{Landwehr2005,Sumner2005}, which are classification trees with logistic regression functions at the leaves.
This is done by using the LogitBoost algorithm.  
In this case, boosting is used (aiming) to build very effective decision trees.
The idea of LMT is to use LogitBoost to induce trees with linear-logistic regression models at the leaves.
LogitBoost performs additive logistic regression. Thus, at each iteration of the boosting algorithm, it creates a simple regression model by going through all
the attributes, finding the simple regression function with the smallest error, and adding it into the additive model \citep{Witten2016}.
The algorithm can deal with binary and multi-class target variables, numeric and nominal attributes and missing values. \textbf{Parameters:}
\begin{itemize}
  \item \underline{Minimum number of objects (\emph{mno})[-M]}: The minimum number of instances per leaf. It can take values in the interval:  \\ $\{mno \in \mathbb{Z} \text{ } | \text{ } 1 \leq mno \leq 64 \}$. \\ \textbf{Default value: }15.
  
  \item \underline{Convert Nominal (\emph{cn})[-B]}: It decides if the algorithm will convert all nominal attributes to binary ones before building the tree. 
  This means that all splits in the final tree will be binary. It can take Boolean values (true or false). \\ \textbf{Default value: }false.
  
   \item \underline{Split on residuals (\emph{sor})[-R]}:  It decides whether the algorithm will set the splitting criterion based on the residuals of LogitBoost. 
   There are two possible splitting criteria for LMT: the default is to use the C4.5 splitting criterion that uses information gain on the class variable. 
   The other splitting criterion tries to improve the purity in the residuals produced when fitting the logistic regression functions. 
   It can take Boolean values (true or false). \\ \textbf{Default value: }false.
    
   \item \underline{Fast Regression (\emph{fr})[-C]}: It decides whether the algorithm will use a heuristic that avoids the use of
   cross-validation to optimize the number of Logit-Boost iterations at every node. 
   In the case of using this heuristic, LMT will fit the logistic regression functions at a leaf node using the LogitBoost
algorithm, applying a 5-fold cross-validation procedure to determine how many iterations to run just once. 
Then, it employs the same number of iterations throughout the tree, instead of cross-validating at every node.   
   This heuristic reduces the running time considerably, with little effect on accuracy \citep{Witten2016}.
     It can take Boolean values (true or false). \\ \textbf{Default value: }true.
   
   \item \underline{Error on probabilities (\emph{eop})[-P]}:  It decides if the algorithm will minimize the error on classification probabilities instead of 
   the misclassification error when cross-validating the number of LogitBoost iterations. 
   When this parameter is set to `true', the number of LogitBoost iterations that minimizes the error on classification probabilities instead of the misclassification error is chosen.
   It can take Boolean values (true or false). \\ \textbf{Default value: }false.
   
   \item \underline{Weight trim beta(\emph{wtb})[-W]}: It sets the beta value used for weight trimming in LogitBoost. 
   Only instances carrying (1 - beta)\% of the weight from the previous iteration are used in the next iteration. 
   The value zero (0) means no weight trimming, which is the default value. 
    The values are restricted to the interval :
   $\{wtb \in \mathbb{R} \text{ } | \text{ } 0.0\leq wtb \leq 1.0 \}$.
   \\ \textbf{Default value: }0.0.
   
    
   \item \underline{Use AIC (\emph{uaic})[-A]}:  It decides if the algorithm will use the AIC (Akaike's Information Criterion) measure to determine when to stop LogitBoost's iterative process. 
    More precisely, if $uaic$ takes the value `true', the best number of iterations will be defined by an information criterion measure (currently, AIC). 
    If false, the stopping criterion will be determined by the best number of iterations in a 5-fold cross-validation procedure.
    It can take Boolean values (true or false).  \\ \textbf{Default value: }false.
\end{itemize}

There are no dependencies/constraints between the parameters of LMT.

\subsection{Decision Stump} \label{slc:ds}

The algorithm for building and applying a decision stump (DS) model \citep{Witten2016}, which is considered a weak learner. 
Because of that, it is usually used in conjunction with a boosting algorithm. 

The DS's classification is based on the entropy measure and a missing value is treated as a separate value.
The DS algorithm constructs a simple decision tree that has only one level, i.e., 
a decision tree that has only one internal (root) node, that is directly linked to the leaves.
It also creates an extra branch for missing values.

In the case of nominal attributes at the root node, there are two possibilities. The first possibility is to build a stump which contains
a leaf for each possible feature value. The second possibility is to consider a stump with two leaves,
one of them is mapped to some category, and the another to all other categories. The DS from WEKA employs the latter approach.
This algorithm has no explicit parameters.


\subsection{Random Forest}\label{slc:rf}

The Random Forest (RF) algorithm for constructing a forest of random trees \citep{Breiman2001}. \textbf{Parameters:}
\begin{itemize}
 \item \underline{Number of trees (\emph{nt})[-I]}:  The number of trees to be generated by the algorithm. It is an integer value bounded by the interval: 
 $\{nt \in \mathbb{Z} \text{ } | \text{ } 2 \leq nt \leq 256\}$.\\ \textbf{Default value: }100.
 \item \underline{Number of features (\emph{nf})[-K]}:  It sets the number of randomly sampled attributes used as candidate attributes at each tree node. It is an integer value bounded by the interval: 
 $\{nf \in \mathbb{Z} \text{ } | \text{ } 2 \leq nf \leq 32\}$. 
 However, it may also take the value zero (0), which means $nf$ will be just used as a flag to indicate that the real value produced by the equation \\
 $log_2 (number\_of\_attributes + 1)$ rounded to the nearest integer is automatically used for this parameter. 
 \\ \textbf{Default value: }0.
 
 \item \underline{Maximum depth (\emph{md})[-depth]}: The maximum depth of the tree. It is bounded by the interval: 
 $\{md \in \mathbb{Z} \text{ } | \text{ } 2 \leq md \leq 20 \}$. 
 However, it may also take the value zero (0) as a flag and, in this case, the depth of the tree can be unlimited.
 \\ \textbf{Default value: }0.
 
\end{itemize}

There are no dependencies/constraints between the parameters of RF.

 \subsection{Random Tree}\label{slc:rt}
 
 The algorithm for constructing a tree that considers K randomly sampled attributes as candidate attributes at each node, i.e., a random tree (RT)~\citep{Witten2016}. It is important to mention that this version of RT performs no pruning. \textbf{Parameters:}
 
\begin{itemize}
  \item \underline{Minimum weight (\emph{mw})[-M]}: The minimum total weight of the instances in a leaf. It is restricted by the interval:  \\ 
  $\{mw \in \mathbb{Z} \text{ } | \text{ } 1 \leq mw \leq 64 \}$.  \\ \textbf{Default value: }1.
 \item \underline{Number of features (\emph{nf})[-K]}:  It sets the number of randomly sampled attributes used as candidate attributes at each tree node. It is an integer value bounded by the interval: 
 $\{nf \in \mathbb{Z} \text{ } | \text{ } 2 \leq nf \leq 32\}$. 
 However, it may also take the value zero (0), which means $nf$ will be just used as a flag to indicate that the real value produced by the equation \\
 $log_2 (number\_of\_attributes + 1)$ rounded to the nearest integer is automatically used for this parameter. 
 \\ \textbf{Default value: }0.
 
 \item \underline{Maximum depth (\emph{md})[-depth]}: The maximum depth of the tree. It is bounded by the interval: 
 $\{md \in \mathbb{Z} \text{ } | \text{ } 2 \leq md \leq 20 \}$. 
 However, it may also take the value zero (0) as a flag and, in this case, the depth of the tree can be unlimited.
 \\ \textbf{Default value: }0.
 
 \item \underline{Number of folds for back-fitting and for growing the tree}\\ \underline{(\emph{nfbgt})[-N]}: 
 It determines the amount of data used for back-fitting and for growing the tree. 
 One fold is used for back-fitting, i.e., for making a preliminary estimation of class probabilities based on a hold-out set.
 The others  ($nf$  - 1) folds are used for growing the tree.
 It is bounded by the interval:   $\{nfbf \in \mathbb{Z} \text{ } | \text{ } 2 \leq nfbf \leq 5 \}$. 
 It can also use the value zero (0), which means no back-fitting will be performed in this case.
 It can not take the value one (1) because we would have zero folds for growing the tree. In the case of taking the value one, the algorithm returns an error
 and does not run. It is important to mention that Auto-WEKA allows this error, ignoring RT algorithm with this configuration (when it occurs), and continuing the search from this  point.
  \\ \textbf{Default value: }0.

 
 \end{itemize}
 
There is no constraints/dependencies between the parameters of RT.

\subsection{REPTree}\label{slc:reptree}

The algorithm for the fast decision tree learner, which is well-known as REPTree~\citep{Witten2016}.
It builds a decision tree using information gain and prunes it using reduced-error pruning (with back-fitting).  
It only sorts values for numeric attributes once, at the start of the algorithm. Missing values are dealt with by splitting the corresponding instances into pieces (i.e., as in C4.5).

\textbf{Parameters:}
\begin{itemize}
  \item \underline{Minimum weight (\emph{mw})[-M]}: The minimum total weight of the instances in a leaf. It is restricted by the interval:  \\ 
  $\{mw \in \mathbb{Z} \text{ } | \text{ } 1 \leq mw \leq 64 \}$.
   \\ \textbf{Default value: }2.
  \item \underline{Maximum depth (\emph{md})[-L]}: The maximum tree depth. It can take integer values considering the interval:  
  $\{md \in \mathbb{Z} \text{ } | \text{ } 2 \leq md \leq 20 \}$.  
   However, it may also take the value $-1$  as a flag and, in this case, the depth of the tree the depth will not be restricted.
  \\ \textbf{Default value: }-1.

  \item \underline{Use pruning (\emph{up})[-P]}: 
  It decides whether REPTree will use reduced-error pruning or not. 
  In the case of using this pruning method, a simple hold-out set ($\frac{1}{3}$ of the training data) is used to estimate the error of a node, instead of using cross-validation.
  It can take Boolean values (true or false).
   \\ \textbf{Default value: }false.

\end{itemize}

There are no dependencies/constraints between the parameters of REPTree.


\subsection{Decision Table} \label{slc:dt}

The algorithm for building and using a simple decision table (DT) classifier \citep{Kohavi1995}.
\textbf{Parameters:}
\begin{itemize}
\item \underline{\emph{Evaluation Measure} (em)[-E]}: The measure used to evaluate the performance of attribute combinations used in the decision table.
It can take one of the four categorical values: 
 1. accuracy (acc);
 2. root mean squared error (rmse) of the the class probabilities;
 3. mean absolute error (mae) of the class probabilities;
 4. area under the ROC curve~(auc). The two measures \emph{rmse} and \emph{mae} are adapted to be used in the classification context.
  \\ \textbf{Default value: }acc.

\item \underline{Use IBk (\emph{uibk})[-I]}: It sets whether a k-nearest neighbor (k=1) classifier should be used instead of the majority class
in order to classify non-matching instances.  
It can take Boolean values (true or false).
 \\ \textbf{Default value: }false.

\item \underline{Search method (\emph{sm})[-S]}: It sets the search algorithm which will be used to find good attribute combinations for the decision table.
It can take the values Greedy Stepwise or Best First. 
 \\ \textbf{Default value: }Best First.

\item \underline{Cross-Validation (\emph{crv})[-X]}: It sets the number of folds for the internal cross validation procedure to evaluate the attribute sets. 
It may take the values one (1), two (2), three~(3) or four (4). If the value one (1) is set for this hyper-parameter, a leave one out procedure is applied.
 \\ \textbf{Default value: }1.
\end{itemize}

There are no dependencies/constraints between the parameters of DT.


\subsection{JRip} \label{slc:jrip}

The algorithm that implements a propositional rule learner algorithm, namely Repeated Incremental Pruning to Produce Error Reduction (RIPPER)~\citep{Cohen1995}. \textbf{Parameters:}
\begin{itemize}
 \item \underline{Minimum total weight (\emph{mtw})[-N]}: This parameter determines the minimum total weight of the instances in a rule.
  It can take values considering the interval:  $\{mtw \in \mathbb{R} \text{ } | \text{ } 1.0 \leq mtw \leq 5.0 \}$.
  \\ \textbf{Default value: }2.0.
 \item \underline{Check error rate (\emph{cer})[-E]}: It decides whether JRip will consider the ``error rate greater or equal than 0.5'' as a stopping criterion. 
 It can take Boolean values (true or false).
 \\ \textbf{Default value: }true.
 \item \underline{Use pruning (\emph{up})[-P]}: It decides whether JRip will use reduced error pruning or not. 
 In the case of using this pruning method, a 3-fold cross-validation procedure is applied to prune the rules.
 Otherwise, no pruning method is used.
 It can take Boolean values (true or false).
 \\ \textbf{Default value: }false.
 \item \underline{Optimizations (\emph{o})[-O]}:  The number of optimization runs.  It can take integer values considering the interval:  $\{o \in \mathbb{Z} \text{ } | \text{ } 1 \leq o \leq 5 \}$.
 \\ \textbf{Default value: }2.
\end{itemize}

There are no dependencies/constraints between the parameters of JRip.

\subsection{One Rule} \label{slc:1R}

The algorithm for building and using a one rule (OneR) classifier \cite{Holte1993}. In other words, it uses the minimum-error attribute for prediction, discretizing numeric attributes. \textbf{Parameters:}
\begin{itemize}
 \item \underline{Minimum bucket size (\emph{mbs}})[-B]: It is used for discretizing numeric attributes. It is limited by the interval: 
 $\{mbz \in \mathbb{Z} \text{ } | \text{ } 1 \leq mbz \leq 32 \}$.
 \\ \textbf{Default value: }6.
  
\end{itemize}

OneR has only one parameter and, consequently, there is no dependencies/constraints for it.

\subsection{PART} \label{slc:part}

The algorithm for generating a PART decision list~\citep{Frank1998}. PART uses the separate-and-conquer paradigm: 
It builds a partial C4.5 decision tree in each iteration and makes the ``best'' leaf 
into a rule.
\textbf{Parameters:}
\begin{itemize}
\item \underline{Minimum number of objects (\emph{mno})[-M]}: The minimum number of instances per leaf. It can take values in the interval:  \\ $\{mno \in \mathbb{Z} \text{ } | \text{ } 1 \leq mno \leq 64 \}$.
\\ \textbf{Default value: }2.
  \item \underline{Binary splits (\emph{bs})[-B]}: It decides whether C4.5 will use binary splits on nominal attributes when building the trees.  It can take Boolean values (true or false).
  \\ \textbf{Default value: }false.
  \item \underline{Reduced-error pruning(\emph{rep})[-R]}: It is used to decide whether reduced-error pruning is used instead of C4.5's default pruning (error-based pruning). 
  If C4.5's error-based pruning is chosen, a (default) confidence factor of 0.25 is used to prune the tree.
  If not (i.e, the reduced-error pruning is chosen), the algorithm will consider each node for pruning and the removal of a subtree at a node is done if the resulting tree performs no worse than the original one on the validation set.
  The size of the validation set is determined by the next parameter ($nr$).
  It can take Boolean values (true or false).
  \\ \textbf{Default value: }true.
 \item  \underline{Number of folds (\emph{nr})[-N]}: It determines the amount of data used for reduced-error pruning.  One fold is used for pruning and the rest for growing the tree.
  It can take the values two (2), three (3), four (4) or five (5).
  \\ \textbf{Default value: }not used.
 \end{itemize}

 \textbf{Dependencies/Constraints:}
 \begin{enumerate}
  \item If the reduced-error pruning method is not set to ``true'', the parameter ``number of folds'' is not used.
 \end{enumerate}
 


\subsection{Zero Rule} \label{slc:0R}

The algorithm for building and using a zero rule (ZeroR) classifier~\citep{Witten2016}. 
The ZeroR classifier simply predicts the majority category (class), ignoring the predictor attributes.
This algorithm has no explicit parameters.

\subsection{K-Nearest Neighbors} \label{slc:knn}

The algorithm for k-nearest neighbors (KNN) classifier~\citep{Aha1991}. KNN can select an appropriate value of K based on internal leave-one-out evaluation and can also compute distances based on instance weighting. \textbf{Parameters:}
\begin{itemize}
\item \underline{Number of neighbors (\emph{k})[-K]}: The number of neighbors to use. The value of $k$ is bounded by the interval: $\{k \in \mathbb{Z} \text{ } | \text{ } 1 \leq k \leq 64 \}$.
\\ \textbf{Default value: }1.
\item \underline{Leave-one-out (\emph{loo})[-X]}: It decides whether leave-one-out evaluation on the training data will be used or not to select the best k value between 1 and the value specified 
as the KNN parameter. If set as false, the selected k value is used. It can take only Boolean values (true or false).
\\ \textbf{Default value: }false.
\item \underline{Distance weighting (\emph{dw})}: It sets the used distance weighting method. It may take the unique following values:
\begin{itemize}
 \item -I: Weight neighbors by the inverse of their distance.
 \item -F: Weight neighbors by one minus their distance.
 \item None: No distance weighting method is applied.
\end{itemize}
\textbf{Default value: }None.

\end{itemize}

There are no dependencies/constraints between the parameters of KNN.

\subsection{K*} \label{slc:k*}

The algorithm K* is an instance-based classification algorithm~\citep{Cleary1995}. Thus, in order to classify a test instance, K* considers 
the class of those training instances similar to it, 
as determined by some similarity function.  It differs from other instance-based learners by using an entropy-based distance function.
\textbf{Parameters:}
\begin{itemize}
\item \underline{Global blending (\emph{gb})[-B]}: The parameter is a percentage for global blending.  This parameter
controls the ``sphere of influence'' by specifying how many of the neighbors of the instance $i$ should be considered
important (although there is no hard cut off at the edge of the sphere -- it is more related to a gradual decreasing of importance).
The values are restricted to the interval 
 $\{gb \in \mathbb{Z} \text{ } | \text{ } 1 \leq gb \leq 100\}$.
Thus, selecting zero (0) for this parameter gives a nearest neighbor algorithm (this is why Auto-WEKA does not allow to choose it), 
and choosing 100 gives equally weighted instances. 
Intermediate values are interpolated linearly.
\\ \textbf{Default value: }20.

\item \underline{Entropic auto-blending (\emph{eab})[-E]}: It decides whether entropy-based blending will be used or not. It can take Boolean values (true or false).
\\ \textbf{Default value: }false.

\item \underline{Missing Mode  (\emph{mm})[-M]}: It determines how missing attribute values are treated. 
It can take one of the four categorical values:
1. average column entropy curves (a);
2. ignore the instances with missing values (d);
3. treat missing values as maximally different (m); 
4. normalize over the attributes (n).
\\ \textbf{Default value: }a.
\end{itemize}

There are no dependencies/constraints between the parameters of K*.



\subsection{Voted Perceptron} \label{slc:vt}

The voted perceptron (VP) algorithm created by \cite{Freund1999}. 
It globally replaces all missing values by their default values. More precisely,
VP replaces all missing values for nominal and numeric attributes by the modes and the means from the training data, respectively.
Additionally, it transforms nominal attributes into binary ones. 
\textbf{Parameters:}
\begin{itemize}
  \item \underline{Number of iterations (\emph{i})[-I]}:  The number of iterations to be performed by VP.  This parameter varies in accordance to the interval: \\ $\{i \in \mathbb{Z} \text{ } | \text{ } 1 \leq i \leq 10 \}$.
  \\ \textbf{Default value: }1.
  \item \underline{Max K(\emph{mk})[-M]}:  The maximum number of alterations to the perceptron, i.e., the maximum number of perceptrons used in the iterative process. It can take values of the interval:  \\ $\{mk \in \mathbb{Z} \text{ } | \text{ } 5,000 \leq mk \leq 50,000 \}$
  \\ \textbf{Default value: }1,000.
  \item \underline{Exponent (\emph{e})[-E]}:  The exponent for the polynomial kernel. It can take values of the interval:   $\{e \in \mathbb{R} \text{ } | \text{ } 0.2\leq e \leq 5.0 \}$
  \\ \textbf{Default value: }1.0.
\end{itemize}

There are no dependencies/constraints between the parameters of VP.


\subsection{Multi-Layer Perceptron} \label{slc:mlp}

Multi-Layer Perceptron (MLP) uses the traditional back-propagation algorithm  \citep{Rumelhart1986} to create a neural model to classify the instances.
MLP creates just one hidden layer (for now) and all its nodes
use sigmoid activation functions (except for when the class is numeric in which case the the output nodes become unthresholded linear units).
\textbf{Parameters:}
\begin{itemize}
 \item \underline{Learning rate (\emph{lrt})[-L]}: The amount by which the weights are updated during training. It is restricted by the interval: \\ $\{lr \in \mathbb{R} \text{ } | \text{ } 0.1 \leq lr \leq 1.0 \}$.
 \\ \textbf{Default value: }0.3.
  \item \underline{Momentum (\emph{m})[-M]}: It is applied to the weights during updating. It is restricted by the interval: $\{m \in \mathbb{R} \text{ } | \text{ } 0.0 \leq m \leq 1.0 \}$.
  \\ \textbf{Default value: }0.2.
  \item \underline{Number of hidden nodes (\emph{nhn})[-H]}: 
  It defines the number of hidden nodes in the hidden layer of the neural network. This parameter may take four predefined nominal values (a, i, o and t), which represent
  the following integer values:
  \begin{itemize}
   \item a = $ \frac{(number\_of\_attributes + number\_of\_classes)}{2} $, always using the default floor function to convert it to an integer value. 
   \item i = number of attributes.
   \item o = number of classes.
   \item t = $(number\_of\_attributes + number\_of\_classes)$.   
  \end{itemize}
  \textbf{Default value: }a.
  
  \item \underline{Nominal to binary filter (\emph{n2b})[-B]}: It decides whether the algorithm will transform nominal attributes to binary ones or not.
  This could help improve performance if there are nominal attributes in the data. It can take Boolean values (true or false).
  \\ \textbf{Default value: }true.
 

  \item \underline{Reset (\emph{r})[-R]}:  It decides whether the algorithm will use the reset approach. In this case, the algorithm will allow the network 
  to reset with a lower learning rate. If the network diverges from the answer, this will automatically reset the network with a lower learning rate and begin training again. 
  It can take Boolean values (true or false).
  \\ \textbf{Default value: }true.
  
  \item \underline{Decay (\emph{d})[-D]}: It decides whether the algorithm will cause the learning rate to decrease. 
  This will divide the starting value of the learning rate by the sequential number of the current epoch in order to determine what the current learning rate should be. 
  This may help to stop the network from diverging from the target output, as well as improving general performance. 
 It can take Boolean values (true or false).
 \\ \textbf{Default value: }false.
\end{itemize}

There are no dependencies/constraints between the parameters of MLP.


\subsection{Stochastic Gradient Descent} \label{slc:sgd}

The algorithm that implements the stochastic gradient descent (SGD) approach \citep{Witten2016} for learning various linear models 
(binary class SVM, binary class logistic regression, squared loss, Huber loss and epsilon-insensitive loss linear regression). 
\textbf{Parameters:}
\begin{itemize}
  \item \underline{Loss function (\emph{lf})[-F]}:  It sets the loss function to be minimized. It can take the following integer values associated to three approaches:
  \begin{itemize}
   \item (0): hinge loss (SVM).
   \item (1): log loss (logistic regression).
   \item (2): squared loss (regression).
  \end{itemize}
\textbf{Default value: }0.
  
  \item \underline{Learning rate (\emph{lrt})[-L]}: The learning rate. 
  If normalization is turned off, then the default learning rate will need to be reduced.
  It is restricted by the interval:  $\{lr \in \mathbb{R} \text{ } | \text{ } 0.00001 \leq lr \leq 1.0 \}$. 
  \\ \textbf{Default value: }0.01.
  
  \item \underline{Ridge (\emph{r})[-R]}: It sets the Ridge value in the log-likelihood. This parameter can take any value of the given set: \\ 
  $\{r \in \mathbb{R} \text{ } | \text{ } 10^{-12}\leq r \leq 10.0 \}$
  \\ \textbf{Default value: }0.0001

  \item \underline{Do not normalize (\emph{nn})[-N]}: It decides whether normalization will be turned off or not. It can take Boolean values (true or false).
  \\ \textbf{Default value: }false.
  
  \item \underline{Do not replace missing values (\emph{nrmv})[-M]}: 
  It decides whether global replacement of missing values will be turned off or not. 
  In the case of being turned off, the missing values will be ignored. 
  Otherwise, SGD will replace all missing values for nominal and numeric attributes by the modes and the means from the training data, respectively.
  It can take Boolean values (true or false).
  \\ \textbf{Default value: }false.
\end{itemize}

There are no dependencies/constraints between the parameters of SGD.

\subsection{Sequential Minimal Optimization} \label{slc:smo}
This algorithm implements John Platt's sequential minimal optimization (SMO) algorithm for training a support vector classifier (SVC) \citep{Platt1999,Keerthi2001,Hastie1998}. It globally replaces all missing values by their default values. More precisely,
SMO (like VP) replaces all missing values for nominal and numeric attributes by the modes and the means from the training data, respectively.
Additionally, it transforms nominal attributes into binary ones. 
\textbf{Parameters:}
\begin{itemize}

\item \underline{Cost (\emph{c})[-C]}: It defines the complexity parameter, 
which is the penalty parameter of the error term and is defined by the interval:  $\{c \in \mathbb{R} \text{ } | \text{ } 0.5 \leq c \leq 1.5 \}$.
This is a parameter that controls the trade-off between training error and model complexity. 
It is important to mention that a low value of $c$ will increase the number of training errors, whereas a high value of $c$ will lead to 
a behavior similar to that of a hard-margin SVM \citep{Joachims2002}.
\\ \textbf{Default value: }1.0.

\item \underline{Filter type (\emph{ft})[-N]}: It determines how/if the data will be transformed.
It may take the values 
zero (0, i.e, normalize the training data -- it sets all the numeric attributes in the given dataset into the interval [0,1]), 
one (1, i.e, standardize the training data --  it standardizes all numeric attributes in the given dataset to have zero mean and unit variance) or  
two (2, i.e. no normalization/standardization is applied to the data).
\\ \textbf{Default value: }0.

 \item \underline{Build Calibration Models (\emph{bcm})[-M]}: It decides whether the model will fit calibration models to SVM's outputs (for proper probability estimates).
 It can take Boolean values (true or false).
 \\ \textbf{Default value: }false.

\item \underline{Kernel(\emph{k})[-K]}: The kernel to use. It can take one of the following possible kernels (and associated constrained parameters):
\begin{itemize}

 \item \underline{PolyKernel}: The standard polynomial kernel. \textbf{Parameters:}
 \begin{enumerate}
  \item \underline{Exponent (\emph{exp})[-E]}: It determines the exponent value and is defined by the interval:  $\{exp \in \mathbb{R} \text{ } | \text{ } 0.2 \leq exp \leq 5.0 \}$.
  \item \underline{Use Lower-Order (\emph{ulo})[-L]}:  It decides whether the algorithm will use lower-order terms or not. It can take Boolean values (true or false).
 \end{enumerate} 

 \item \underline{NormalizedPolyKernel}: The normalized polynomial kernel. \textbf{Parameters:}
 \begin{enumerate}
  \item \underline{Exponent (\emph{exp})[-E]}: It determines the exponent value and is defined by the interval:  $\{exp \in \mathbb{R} \text{ } | \text{ } 0.2 \leq exp \leq 5.0 \}$.
  \item \underline{Use Lower-Order (\emph{ulo})[-L]}:  It decides whether the algorithm will use lower-order terms or not. It can take Boolean values (true or false).
 \end{enumerate}

 \item \underline{Puk}: The Pearson VII function-based universal kernel \citep{Ustun2006}. \textbf{Parameters:}
 \begin{enumerate}
  \item \underline{Omega (\emph{om})[-O]}: The omega value. It is defined by the interval: $\{om \in \mathbb{R} \text{ } | \text{ } 0.1\leq om \leq 1.0 \}$.
  \item \underline{Sigma (\emph{sig})[-S]}: The sigma value. It is defined by the interval: $\{sig \in \mathbb{R} \text{ } | \text{ } 0.1 \leq sig \leq 10.0 \}$.
 \end{enumerate}  
 
  \item \underline{RBF}:The RBF kernel. \textbf{Parameters:}
 \begin{enumerate}
  \item \underline{Gamma (\emph{g})[-G]}: The gamma value. It is defined by the interval: $\{g \in \mathbb{R} \text{ } | \text{ } 0.0001\leq g \leq 1.0 \}$.
 \end{enumerate}   
 
\end{itemize}
\textbf{Default value: } PolyKernel with `Exponent' equals to 1.0 and `Use Lower-Order' equals to true.
\end{itemize}

The constraints/dependencies for SMO are only in the selection of the kernel and its respective parameters.

\subsection{Logistic Regression} \label{slc:lr}

The algorithm for building and using a multinomial logistic regression (LogR) model with a ridge estimator \citep{Cessie1992}. 
\textbf{Parameters:}
\begin{itemize}
\item \underline{Ridge (\emph{r})[-R]}: It sets the Ridge value in the log-likelihood. This parameter can take any value of the given set: \\ 
$\{r \in \mathbb{R} \text{ } | \text{ } 10^{-12}\leq r \leq 10.0 \}$
\\ \textbf{Default value: } 0.00000001
\end{itemize}

LogR has one parameter and, consequently, there is no dependencies/constraints for it.


\subsection{Simple Logistic} \label{slc:sl}

The algorithm for constructing simple logistic regression (SL) models  \citep{Landwehr2005,Sumner2005}. LogitBoost with simple regression functions as base learners is used for fitting the logistic models.
\textbf{Parameters:}
\begin{itemize}
\item \underline{Weight trim beta (\emph{wtb})[-W]}: It sets the beta value used for weight trimming in LogitBoost. 
   Only instances carrying (1 - beta)\% of the weight from the previous iteration are used in the next iteration. The value zero (0) means no weight trimming, which is the default value. 
    The values are restricted to the interval :
   $\{wtb \in \mathbb{R} \text{ } | \text{ } 0.0\leq wtb \leq 1.0 \}$. It also can be omitted and take the default value of zero. 
   \\ \textbf{Default value: }0.0.
   
   \item \underline{Use Cross-Validation (\emph{ucv})[-S]}: 
   It decides if SL will try to find the best number of LogitBoost iterations using an internal 5-fold cross-validation procedure or simply using the 
   number of iterations that minimizes error on the training set.   
 Thus, if not set to `true', the number of LogitBoost iterations which is used is the one that minimizes the error on the training set (misclassification error).  It can take Boolean values (true or false).
 \\ \textbf{Default value: }true.

    \item \underline{Use AIC (\emph{uaic})[-A]}:  It decides if the method will use the AIC (Akaike's Information Criterion) measure to determine when to stop the LogitBoost iterative process. 
    More precisely, if $uaic$ takes the value 'true', the best number of iterations will be defined by an information criterion measure (currently, AIC).
    If false, the stopping criterion will be determined by the best number of iterations in an internal 5-fold cross-validation procedure or simply in accordance to the error on the training set, as explained in the previous item.
    It can take Boolean values (true or false).
    \\ \textbf{Default value: }false.

\end{itemize}

There are no dependencies/constraints between the parameters of SL.
\subsection{Na\"ive Bayes}\label{slc:nb}

The Na\"ive Bayes (NB) classifier using estimator classes  \citep{John1995}. This algorithm builds a fixed structure (model) given the attributes of the dataset. \textbf{Parameters:}
\begin{itemize}
  \item \underline{Use kernel estimator (\emph{uke})[-K]}: 
  It decides whether NB will use a kernel estimator for numeric attributes rather than a (single) Gaussian distribution.  
  In the case of using the kernel estimator, NB will apply one Gaussian kernel per observed data value (for more details, see Flexible Na\"ive Bayes' section in \cite{John1995}).
  It can take Boolean values (true or false). 
  It is important to mention that a discrete estimator is automatically used for nominal attributes,
  which is a simple discrete probability estimator based on nominal values' counts. 
  This also means the Laplace correction is applied in order to perform the estimation.
  \\ \textbf{Default value: }false.
  
  \item \underline{Use supervised distribution (\emph{usd})[-D]}: It decides whether NB will use supervised discretization to convert numeric attributes to nominal ones. 
  Discretization is performed by the algorithm proposed in \cite{Fayyad1993}. This method uses a criterion based on the minimum description length (MDL) principle
  to define the number of intervals produced over the continuous space \cite{Dougherty1995}.
  It can take Boolean values (true or false).
  \\ \textbf{Default value: }false.
\end{itemize}

 \textbf{Dependencies/Constraints:}
 \begin{enumerate}
  \item If the parameter ``use kernel estimator'' is activated, the parameter ``use supervised distribution'' must not be activated; and vice-versa.
  This constraint must be enforced in the grammar.
 \end{enumerate}

 \subsection{Bayesian Network Classifier}\label{slc:bnc}
This algorithm is used to learn a Bayesian Network Classifier (BNC) \citep{Bouckaert2007} based on various search algorithms and a local Bayesian scoring metric \citep{Cooper1992,Heckerman1994}. 
\textbf{Parameters:}
\begin{itemize}
\item \underline{Search Method (\emph{sm})[-Q]}:  For BNC algorithms, the optimization occurs just on the method used for searching network structures. 
Thus, the search method  can be one of the following: 
1. Tree Augmented Na\"ive Bayes (TAN) \citep{Friedman1997};
2. K2 \citep{Cooper1992};
3. Hill Climbing (HC) \citep{Bouckaert1995, Hesar2012};
4. Look Ahead in Good Directions Hill Climbing (LAGDHC) \citep{Abramovici2008};
5. Simulated Annealing (SA) \citep{Bouckaert1995};
6. Tabu Search (TS) \citep{Bouckaert1995}. 
All the search methods uses the parameter ``maximum number of parents'' set to two (including the class node), except for TAN and SA which do not use the 
``maximum number of parents'' as a parameter.
In addition, the methods use the (default) Bayesian scoring metric to search for appropriate Bayesian networks to data. 
\\ \textbf{Default value: }there is not default value for this hyper-parameter because all these search methods are important algorithms in the literature. Therefore, we use all search methods in our space.
\end{itemize}

BNC has one parameter and, consequently, there is no dependencies/constraints for it.


\subsection{Na\"ive Bayes Multinomial}\label{slc:nbm}

The algorithm for building and using Na\"ive Bayes Multinomial (NBM) \citep{Aggarwal2012,Frank2006,Lewis1998,McCallum1998,Witten2016}. This algorithm was particularly designed for text classification and, for this reason,
it changes how the traditional Na\"ive Bayes calculates the probabilities. This is done to take into account the number of times a word appears in the document.

This algorithm has no explicit parameters and, consequently, there is no dependencies/constraints for it.

\section{Search Space -- Meta classification algorithms from WEKA}
\label{meta-slc}

In this section, the \emph{search space} of \textbf{5} traditional (single label) meta classification algorithms from WEKA \citep{Hall2009} is studied. This is also done in order to extend and improve the \emph{search space} of multi-label methods.
All parameters in this section were also set in accordance to the \emph{search space} definition from Auto-WEKA \citep{Thornton2013,Kotthoff2017,Kotthoff2017b}.
The methods and their respective (hyper-)parameters were defined after studying the code, logs and configuration files of Auto-WEKA, which is considered a stable and robust approach for
automatically selecting and configuring machine learning algorithms.

\subsection{Locally Weighted Learning} \label{slc:lwl}

The locally weighted learning (LWL) method \citep{Frank2002,Atkeson1997} . It uses an instance-based algorithm to assign instance weights which are then used by a specified Weighted Instances Handler.
In other words, LWL assigns weights using an instance-based method and, after this step, another classification algorithm is used to build a classifier from the weighted instances.
For example, it can do the classification by using a na\"ive Bayes classifier or a decision stump (default) from these weighted instances.  \textbf{Parameters:}
\begin{itemize}
  \item \underline{Classifier (\emph{c})[-W]}: The classifier to be used. It can be one classification algorithm from Section~\ref{slc},
  except for the algorithms LMT, OneR, K*, SGD and VP, as these the  classifiers produced by these algorithms do not handle weighted instances.
  \item \underline{Number of neighbors (\emph{k})[-K]}: It sets how many neighbors are used to determine the width of the weighting function. 
  It may take the following values: $\{-1, 10, 30, 60, 90, 120\}$. A negative value means that all neighbors will be considered.  
   \\\textbf{Default value: }-1.
  
  \item \underline{Weighting kernel (\emph{wk})[-U]}:  It determines the weighting function and may take the five following integer values:
  \begin{itemize}
   \item (0) Linear.
   \item (1) Epnechnikov.
   \item (2) Tricube.
   \item (3) Inverse.
   \item (4) Gaussian.
  \end{itemize}
It can be omitted with 50\% of probability, and then LWL will use the default value zero for this parameter, i.e., the linear function. 
  The five (5) not omitted values jointly take the other 50\% of probability, which represents at the end that the value $0$ has 60\% of probability to be chosen.
  \\ \textbf{Default value: }0.
\end{itemize}

There are no dependencies/constraints between the parameters of LWL.


\subsection{Random Subspace} \label{slc:rss}

This method constructs an ensemble classifier that consists of multiple models systematically constructed by randomly selecting subsets  of components of the feature vector, 
i.e., the classification models are constructed according to random subspaces (RSS)  \citep{Ho1998}. More precisely,
for each classifier, a certain percentage of the number of attributes is randomly sampled and then used to build the classifier.
\textbf{Parameters:}
\begin{itemize}
   \item \underline{Classifier (\emph{c})[-W]}: The classifier to be used. It can be any classification algorithm from Section~\ref{slc}.
 \item \underline{Subspace size (\emph{sss})[-P]}: It defines the size of each sub-space as a percentage of the number of attributes.
 It could take values in the range: $\{sss \in \mathbb{R} \text{ } | \text{ } 0.1 \leq sss \leq 1.0 \}$.  
  \\ \textbf{Default value: }0.5.
  
   \item \underline{Number of iterations (\emph{ni})[-I]}: It defines the number of iterations to be performed, i.e., the number of classifiers in the ensemble.
 It may take values in the range: \\ $\{ni \in \mathbb{Z} \text{ } | \text{ } 2 \leq ni \leq 64 \}$. 
  \\ \textbf{Default value: }10.
 
\end{itemize}

There are no dependencies/constraints between the parameters of RSS.

\subsection{Bagging of Single-Label Classifiers}\label{slc:bagging}

The method for bagging a classifier in order to reduce variance \citep{Breiman1996}.
\textbf{Parameters:}
\begin{itemize}
 \item \underline{Classifier (\emph{c})[-W]}: The classifier to be used for each member of the ensemble. It can be any classification algorithm from Section~\ref{slc}.
 \item \underline{Bag size percent (\emph{bsp})[-P]}: It defines the size of each bag, as a percentage of the training set size. 
 It may take values in the range: $\{bsp \in \mathbb{Z} \text{ } | \text{ } 10 \leq bsp \leq 100 \}$.
  It makes sampling with replacement. Thus, even if the bag size percent is 100\%, it will sample different sets with the same size of the training set.
  \\ \textbf{Default value: }100.
  
  \item \underline{Number of iterations (\emph{ni})[-I]}: It defines the number of iterations to be performed, i.e., the number of classifiers in the ensemble.
 It may take values in the range: \\ $\{ni \in \mathbb{Z} \text{ } | \text{ } 2 \leq ni \leq 128 \}$. 
 \\  \textbf{Default value: }10.
 
  \item \underline{Calculate out-of-bag  (\emph{coob})[-O]}: It decides whether the out-of-bag error is calculated.
 It can take Boolean values (true or false).
  \\ \textbf{Default value: }false.
\end{itemize}

\textbf{Dependencies/Constraints:}
\begin{enumerate}
 \item If the parameter ``calculate out-of-bag'' is activated (set to true), the parameter ``bag size percent'' must be equal to 100.
 This is a constraint of WEKA and only an internal modification in the WEKA code could suppress it.
 This can happen with 50\% of probability. I.e., in half of the cases, the parameter ``bag size percent'' is set to 100.
 In the other part of the cases, ``bag size percent'' may take values between 10 and 100, because the parameter ``calculate out-of-bag'' is not activated (set to false).
\end{enumerate}


\subsection{Random Committee}\label{slc:rc}

The method for building an ensemble of randomizable base classifiers from WEKA \citep{Witten2016}, creating a random committee (RC) of classifiers. 
For this reason, the only classifiers (at the base level) that can be used in for this meta-algorithm are
Random Forest (RF), Random Tree~(RT), REP Tree (REPTree), Stochastic Gradient Descent (SGD) and Multilayer Perceptron (MLP). 
The creation of a randomizable classifier is done by using an input (pseudo-random) seed.
It is important to mention that the classifiers in the ensemble differ in terms the structure of their models.
For instance, a random seed can define how the random trees are constructed in RF, RT and REPTree, how the linear models are defined in SGD, and
how the network connection weights are firstly defined in MLP. Nevertheless, all classifiers are constructed using the same data, differently from
Bagging and RSS.
Thus, at the end, the final prediction is based on the average of the class probabilities generated by the base classifiers. \textbf{Parameters:}
\begin{itemize}
     \item \underline{Classifier (\emph{c})[-W]}: The classifier to be used for each member of the ensemble. 
     It is restricted in one of the five (5) algorithms aforementioned, i.e., RF, RT, REPTree, SGD and MLP.
     

   \item \underline{Number of iterations (\emph{ni})[-I]}: It defines the number of iterations to be performed, i.e., the number of classifiers in the ensemble.
 It may take values in the range: $\{ni \in \mathbb{Z} \text{ } | \text{ } 2 \leq ni \leq 64 \}$. 
  \\ \textbf{Default value: }10.
\end{itemize}

There are no dependencies/constraints between the parameters of RC.

\subsection{Ada Boost M1}\label{slc:ada}

The method for boosting a nominal class classifier using the Adaboost M1 (AdaM1) approach \citep{Freund1996}. \textbf{Parameters:}
\begin{itemize}
  \item \underline{Classifier (\emph{c})[-W]}: The classifier to be used. It can be one classification algorithm from Section~\ref{slc},
  except for the algorithms LMT, OneR, K*, SGD and VP, as these the  classifiers produced by these algorithms do not handle weighted instances.
  It is important to mention than Auto-WEKA allows any classifier at the base level of AdaM1, including those which can not handle weights in the instances.
  In this case, Auto-WEKA ignores that algorithm (with its configuration) and proceeds with the search.
 \item \underline{Weight threshold (\emph{wt})[-P]}: It defines the weight threshold for weighted pruning, i.e.,
  it only selects instances with weights that contribute to the specified quantile of the weight distribution.  
 It may take values in the range: \\ $\{wt \in \mathbb{Z} \text{ } | \text{ } 50 \leq wt \leq 100 \}$. 
  \\ \textbf{Default value: }100.
 
  \item \underline{Number of iterations (\emph{ni})[-I]}: It defines the number of iterations to be performed, i.e., the number of classifiers in the ensemble.
 It may take the values in the range: \\ $\{ni \in \mathbb{Z} \text{ } | \text{ } 2 \leq ni \leq 128 \}$. 
  \\ \textbf{Default value: }10.
 
 \item \underline{Use resampling (\emph{ur})[-Q]}:  It decides whether AdaM1 will use resampling instead of reweighting.  
 Thus, it is possible to generate an unweighted dataset from the weighted data by resampling. In this case,
 instances are chosen with probability proportional to their weight. As a result, instances with high weight are replicated frequently, and the ones with low
weight may never be selected. Once the new dataset becomes as large as the original one, it is fed into the learning approach instead of the weighted data \citep{Witten2016}.  
 It can take Boolean values (true or false).
 \\ \textbf{Default value: }false.
\end{itemize}

There are no dependencies/constraints between the parameters of AdaM1.

\section{Search Space -- Preprocessing algorithms from WEKA} \label{prep-slc}

In this section, the \emph{search space} of (single label) preprocessing classification algorithms from WEKA \citep{Hall2009} is studied.
This is also done in order to extend and improve the \emph{search space} of multi-label methods.
Instead of using just a single-label classification (SLC) algorithm at the SLC base level, a wrapper containing preprocessing methods is firstly used and, just after that,
SLC is performed.


\subsection{Attribute Selection Classifier} \label{slc:asc}

The method that reduces the dimensionality of training and test data by performing attribute selection (using the training set only)  before the data is set as input to a classifier  \citep{Witten2016}, constructing an attribute selection classifier (ASC). \textbf{Parameters:}
\begin{itemize}
   \item \underline{Classifier (\emph{c})[-W]}: The classifier to be used. 
   It can be any classification algorithm from Section~\ref{slc}.
   
   \item \underline{Search method (\emph{sm})[-S]}: The search method for selecting the attribute subset to be used as input by the classifier. It may take two values:
   \begin{enumerate}
    \item Best First: It searches the space of attribute subsets by greedy hill-climbing augmented with a backtracking facility.
    \item Greedy Stepwise : It performs a greedy forward search through the space of attribute subsets.
   \end{enumerate}
   Both methods use the evaluator ``CfsSubsetEval'', which evaluates the worth of a subset of attributes by considering the individual 
   predictive ability of each attribute along with the degree of redundancy between them.
   Hence, ASC is conceptually equivalent to using the CFS (Correlation-based Feature Selection) attribute selection method followed by the use of the chosen classifier with the attributes selected by CFS.
   \\ \textbf{Default value: }Best First.
\end{itemize}

There are no dependencies/constraints between the parameters of ASC.

%
%
%
%
%


\section{A Formal Description of the MLC Search Space} \label{grammar-mlc}

In this section, we describe the context-free grammar \cite{Sipser2012} used to formally specify the \emph{search spaces} specified in this supplementary material.

Formally, a grammar \emph{G} is represented by a four-tuple \textit{\textless N, T, P, S\textgreater}, where  \emph{N} represents a set of non-terminals,  \emph{T} a set of terminals,  \emph{P} a set of production rules and  \emph{S} (a member of  \emph{N}) the start symbol. In this thesis, we will use the Backus Naur Form (BNF) to represent grammars. This means that each production rule has, for instance, the following form  \textit{\textless Start\textgreater ::= $[$\textless A\textgreater $]$ \textless B\textgreater { $|$} \textless C\textgreater { } $($d $|$ e$)$ }. Symbols wrapped in ``$<\:>$'' represent non-terminals, whereas terminals (such as $d$ and $e$) are not bounded by ``$<\:>$''. The special symbols ``$|$'', ``$[ ]$'' and ``$( )$'' represent, respectively,  a choice, an optional element and a set of grouped elements that should be used together. Additionally, the symbol ``\#'' represents a comment in the grammar, i.e., it is ignored by the grammar's parser. The choice of one among all elements connected by ``$|$'' is made using a uniform probability distribution (i.e., all elements are equally likely to occur in an individual).

The proposed grammar has 125 production rules, in a total of 124 non-terminals and 213 terminals. Figures \ref{fig_gramar1}-\ref{fig_gramar7} present the produced grammar (in the Backus Naur Form) that encompasses the knowledge about multi-label classification in MEKA. This grammar models the \emph{search space} Large, as this \emph{search space} includes all learning algorithms and hyper-parameters from the other \emph{search spaces} (i.e., Small and Medium). To consider the grammar for the other two \emph{search spaces}, it is just needed to exclude the respective production rules, non-terminals and terminals -- which represent the learning algorithms and their respective hyper-parameters -- that do not make part of them. For sake of simplicity, we will not do that here.

In Figure \ref{fig_gramar1}, the first production rule (\text{\textless Start\textgreater }) is used to describe the multi-label classification~(MLC) \emph{search space}. In this grammar rule,  \text{\textless MLC-PT\textgreater} denotes problem transformation,  \text{\textless MLC-AA\textgreater} denotes algorithm adaptation, and  \text{\textless META-MLC-LEVEL\textgreater} denotes the multi-label meta-algorithms. We designed the grammar in such a way that all the 26 MLC algorithms have the same probability of being chosen (i.e., each MLC algorithm has $\approx$3.846\% of chance of being chosen). Furthermore, the MLC algorithms must use a prediction threshold~(\textless pred\_tshd\textgreater), which defines the threshold to perform the classification using the model's confidence outputs \citep{Al2014}. 

\begin{figure*}[!ht]
\fontsize{7pt}{8pt}\selectfont
\centering       
\begin{Verbatim}[frame=single,samepage=true]
<Start> ::= (<MLC-PT> | <MLC-AA> | <META-MLC-LEVEL> ) <pred_tshd>

<pred_tshd> ::= PCut1 | PCutL | RANDOM-REAL(>0.0, <1.0)   #pred_tshd=`prediction threshold'
                                                          #PCut1=`P-Cut method',PCutL=`P-Cut method by Label' 

<MLC-PT> ::= <ALGS-PT> <ALGS-SLC> 

<ALGS-SLC> ::= <ALG-TYPE> | <META1> <ALG-WEIGHTED-TYPE> | <META2> <ALG-RANDOM-TYPE> | <META3> <ALG-TYPE> 

<ALG-TYPE> ::= [ASC <sm>] (<TREES> | <RULES> | <LAZY> | <FUNCTIONS> | <BAYES> | <OTHERS>)
							#ASC=`Attribute Selection Classifier'

<sm> ::= GreedyStepwise | BestFirst                     #sm=`search method'            


<TREES> ::= <C4.5> | DecisionStump | ( ( (RandomForest <nt> | <RandomTree>) <nf> ) | <REPTree> ) <md>  

<C4.5> ::= <C4.5-Basics> ( (<cf> [sr]) | u )            #sr=`subtree raising', u='unpruned'
<C4.5-Basics> ::= <mno> [ct] [bs] [umc] [ul]            #ct=`collapse tree', bs='binary splits'
                                                        #umc=`use MDL correction', ul='use Laplace'
<cf> ::= RANDOM-REAL(0.0, 1.0)                          #cf=`confidence factor'
<mno> ::= RANDOM-INT(1, 64)                             #mmo=`minimum number of objects'

<nt> ::= RANDOM-INT(2, 256)                             #nt=`number of trees'
<nf> ::= RANDOM-INT(2, 32)                              #nf=`number of features'
<md> ::= RANDOM-INT(2, 20)                              #md=`maximum depth'

<RandomTree> ::= <mw> <nfbgt>
<mw> ::= RANDOM-INT(1,64)                               #mw=`minimum weight for instances in a leaf'
<nfbgt> ::= 2 | 3 | 4 | 5                               #nfbgt=`number of folds for back-fitting and 
							#       for growing the tree'                                                         
                                                        
<REPTree> ::= <mw> [up]                                 #up=`use pruning'  
                                                        #mw is not included in the same rule for Random Tree                                                      
                                                        #and for REPTree because of the grammar's constraints
\end{Verbatim}
                                                        \setlength{\fboxsep}{1mm}
    \caption{Defined Grammar -- Part 1: General and SLC Trees Algorithms.}
    \label{fig_gramar1}
\end{figure*}

The grammar rule defining the problem transformation methods, i.e.,  \text{\textless MLC-PT\textgreater}, has two components in the right-hand side, namely the actual problem transformation algorithm \text{\textless ALGS-PT\textgreater} (defined in Figure~\ref{fig_gramar5}) and the single-label classification algorithm (SLC, which is represented by the  rule \text{\textless ALG-SLC\textgreater} in the grammar) to perform the single-label classification task(s). This happens because the problem transformation method transforms the multi-label task into one or more single-label tasks. We start discussing the rule defining \text{\textless ALG-SLC\textgreater}.

We divided the SLC algorithms in six (6) types for the grammar following the WEKA software: 
Trees, Rules, Lazy, Functions, Bayes and Others. The last type was created just to simplify the grammar (i.e., it is not an inherent WEKA's category).
Figure~\ref{fig_gramar1} shows the grammar rules for Tree algorithms.
Figure~\ref{fig_gramar2} shows the grammar rules for Rules and Lazy algorithms.
Figure~\ref{fig_gramar3} shows the grammar rules for the other three types of SLC algorithms.
Figure~\ref{fig_gramar1} also defines the Attribute Selection Classifier (ASC), a wrapper which can be used together with the SCL algorithms.
In this case, a preprocessing method is used before the classification step is performed.

\begin{figure*}[!ht]
\fontsize{7pt}{8pt}\selectfont
\centering     
\begin{Verbatim}[frame=single,samepage=true]
<RULES> ::= <DT> | <JRip> | OneR <mbs> | <PART> | ZeroR 

<DT> ::= <em> [uibk] <sm> <crv>                         #uibk=`use IBk'
							#sm=`search method -- defined earlier'
<em> ::= acc | rmse | mae | auc                         #em=`evaluation measure'
<crv> ::= 1 | 2 | 3 | 4                                 #crv=`number of folds for cross-validation'

<JRip> ::= <mtw> [cer] [up] <o>                         #cer=`check error rate', up=`use pruning'
<mtw> ::= RANDOM-REAL(1.0, 5.0)                         #mtw=`minimum total weight for instances 
							#     covered by a rule'
<o> ::= RANDOM-INT(1,5)                                 #o=`number of optimization runs'

<mbs> ::= RANDOM-INT(1,32)                              #mbs=`minimum bucket size'

<PART> ::= <PART-BASICS> (rep <nr> | ebp)               #rep=`use reduced-error pruning'                                                                            
                                                        #nr=`number of folds for reduced-error pruning'
                                                        #ebp=`use error-based pruning'
<PART-BASICS> ::= <mno> [bs]                            #mno=`minimum number of objects'
<nr> ::= RANDOM-INT(2,5) 

<LAZY> ::= <KNN> | <K*>


<KNN> ::= <k_nn> [loo] [<dw>]                           #loo=`leave-one-out to set the k value given the range'
<k_nn> ::= RANDOM-INT(1,64)                             #k_nn=`number of neighbors'
<dw> ::= F | I                                          #dw=`distance weighting'

<K*> ::= <gb> [eab] <mm>                                #eab=`entropic auto-blending'
<gb> ::= RANDOM-INT(1,100)                              #gb=`global blending'
<mm> ::= a | d | m | n                                  #mm=`missing mode to deal with missing values'     
\end{Verbatim}
    \setlength{\fboxsep}{1mm}
    \caption{Defined Grammar -- Part 2: SLC Rules and Lazy Algorithms .}
    \label{fig_gramar2}
\end{figure*}

\begin{figure*}[!ht]
\fontsize{7pt}{8pt}\selectfont
\centering     
\begin{Verbatim}[frame=single,samepage=true]
<FUNCTIONS> ::= <VotedPerceptron> | <MultiLayerPerc> | 
                (<StocGradDescent> | LogisticRegression) <r> | <SeqMinOptimization>

<VotedPerceptron> ::= <i> <mk> <e>                                   
<i> ::= RANDOM-INT(1,10)                                #i=`number of iterations'
<mk> ::= RANDOM-INT(5000, 50000)                        #mk=`maximum number of alterations to the perceptrons'
<e> ::= RANDOM-REAL(0.2, 5.0)                           #e=`The exponent for the polynomial kernel'


<MultiLayerPerc> ::= <lr> <m> <nhn> [n2b] [r] [d]       #n2b=`nominal to binary filter', 
                                                        #r=`use reset approach', 
                                                        #d=`decay in the learning rate'
<lr> ::= RANDOM-REAL(0.1, 1.0)                          #lr=`learning rate'
<m> ::= RANDOM-REAL(0.0, 1.0)                           #m=`momentum'
<nhn> ::= a | i | o | t                                 #nhl=`rules to define the number of hidden nodes'

<StocGradDescent> ::= <lf> <lr_sgd> [nn] [nrmv]         #nn=`do not normalize', 
                                                        #nrmv=`do not replace missing values'               
<lf> ::= 0 | 1 | 2                                      #lf=`loss function'
<lr_sgd> ::= RANDOM-REAL(0.00001, 1.0)                  #lr_sgd=`learning rate for SGD'
<r> ::= RANDOM-REAL(0.000000000001,10.0)                #r=`ridge value in the log-likelihood'


<SeqMinOptimization> ::= <c> <ft> [bcm] <kernel>        #bcm=`build calibration models'
<c> ::= RANDOM-REAL(0.5,1.5)                            #c=`the cost, i.e.,complexity parameter'
<ft> ::= 0 | 1 | 2                                      #ft=`filter type'
<kernel> ::= ( NormPolyKernel | 
               PolyKernel
             ) <exp> [ulo] |                            #ulo=`use lower order'
              Puk <om> <sig> | RBF <g>
<exp> ::= RANDOM-REAL(0.2, 5.0)                         #exp=`the exponent'
<om> ::= RANDOM-REAL(0.1, 1.0)                          #om=`the omega value'
<sig> ::= RANDOM-REAL(0.1, 10.0)                        #sig=`the sigma value'
<g> ::= RANDOM-REAL(0.001, 1.0)                         #g=`the gamma value'

<BAYES> ::= NaiveBayes [<NB-Parameters>] | <BayesianNetworkClassifiers> | NaiveBayesMultinomial
<NB-Parameters> ::= uke | usd                           #uke=`use kernerl estimator'
                                                        #usd=`use supervised distribution'

<BayesianNetworkClassifiers> ::= TAN | K2 | HillClimber | LAGDHillClimber | SimulatedAnnealing | TabuSearch

<OTHERS> ::= (SimpleLogistic [ucv] |                    #ucv=`use cross-validation'
                  <LogisticModelTrees>
                 ) [uaic] [<wtb>]                       #uaic=`use AIC measure as stopping criteria'
                 
<LogisticModelTrees> ::=  <mno> [cn] [sor] [fr] [eop]   #cn=`convert nominal to binary'
                                                        #sor=`split on residuals'
                                                        #fr=`fast regression', eop=`error on probabilities'
<wtb> ::= RANDOM-REAL(0.0, 1.0)                         #wtb=`weight trim beta'
\end{Verbatim}
    \setlength{\fboxsep}{1mm}
    \caption{Defined Grammar -- Part 3: SLC Functions, Bayes and Others Algorithms.}
    \label{fig_gramar3}
\end{figure*}

It is also important to mention that some methods at the single-label level, such as Decision Stump and ZeroR, do not have user-defined hyper-parameters.
Others, such as the Bayesian Network Classifier algorithms, do not have user-defined hyper-parameters in the Auto-WEKA software, even though they have 
user-defined parameters in WEKA. That is, the developers of Auto-WEKA have chosen to use a fixed predefined number of parameter settings for some algorithms. As we are following Auto-WEKA to define the parameters at this level (because its robustness to select SLC algorithms), the absence of user-defined parameters in some methods was maintained.

At the single-label level, we also have meta-algorithms, divided in three~(3) types: \text{\textless META1\textgreater}, \text{\textless META2\textgreater} and \text{\textless META3\textgreater}. 
These three categories of meta-algorithms are firstly called in Figure \ref{fig_gramar1}. 
As shown in Figure \ref{fig_gramar4}, \text{\textless META1\textgreater} may take the two~(2) meta-algorithms Ada Boost M1 (AdaM1) and Locally Weighted Learning (LWL), which need a base classifier at the SLC level that handles weighted instances. This is the reason the rule
\text{\textless ALG-WEIGHTED-TYPE\textgreater} is defined.
On the other hand, \text{\textless META2\textgreater} may take just one algorithm, i.e., Random Committee. The reason for that is because this SLC
meta-algorithm can only be used with randomizable base classifiers. The rule  \text{\textless ALG-RANDOM-TYPE\textgreater} expresses these randomizable classifiers. 
The least restricted meta-algorithms are Random Subspace and Bagging, which are specified by \text{\textless META3\textgreater}, being able to use any SLC base classifier (from 
\text{\textless ALG-TYPE\textgreater}).

\begin{figure*}[!ht]
\fontsize{7pt}{8pt}\selectfont
\centering      
\begin{Verbatim}[frame=single,samepage=true]            
<META1> ::= <LWL> | <AdaM1>
            
<LWL> ::= <k_lwl> [<wk>]                                #LWL=`Locally Weighted Learning'
<k_lwl> ::= -1 | 10  | 30 | 60 | 90 | 120               #k_lwl=`number of neighbors in LWL'
<wk> ::= 0 | 1 | 2 | 3 | 4                              #wk=`weighting kernel'

<AdaM1> ::= <wt> [ur] <ni_ada_and_bagging>              #ur=`use resampling'
<wt> ::= RANDOM-INT(50, 100) | 100                      #wt=`weight threshold'
<ni_ada_and_bagging> ::= RANDOM-INT(2, 128)             #ni_ada_and_bagging=`number of iterations for 
							#                    AdaM1 and Bagging'


<ALG-WEIGHTED-TYPE> ::= <TREES> | <RULES-PARTIAL> | <KNN> | <BAYES> | <FUNCTIONS-PARTIAL>
<RULES-PARTIAL> ::= <DT> | <JRip> | <PART> | ZeroR
<FUNCTIONS-PARTIAL> ::= <MultiLayerPerc> |  <SeqMinOptimization> | <SimpleLogistic> <uaic> <wtb_activate>



<META2> ::= RandomCommittee <ni_random_methods>
<ni_random_methods> ::= RANDOM-INT(2, 64)               #ni_random_methods=`number of iterations for 
							#random methods'
<ALG-RANDOM-TYPE> ::=  ( ( (RandomForest <nt> | <RandomTree>) <nf> ) | <REPTree> ) <md>  | 
		       <StocGradDescent> <r> | <MultiLayerPerc>


<META3> ::= <Bagging> | <RandomSubspace> 

<Bagging> ::= (<bsp> | 100 coob) <ni_ada_and_bagging>   #coob=`calculate out-of-bag'                    
                                                        #when coob is true, bag percent size must be 100
<bsp> ::= RANDOM-INT(10, 100)                           #bsp=`bag size percent'
<RandomSubspace> ::= <sss> <ni_random_methods> 
<sss> ::= RANDOM-REAL(0.1, 1.0)                         #sss=`subspace size'
\end{Verbatim}
    \setlength{\fboxsep}{1mm}
    \caption{Defined Grammar -- Part 4: SLC Meta-Algorithms.}
    \label{fig_gramar4}
\end{figure*}

It is important to emphasize that all 28 SLC methods (traditional, meta and preprocessing) have the same chance of being chosen by a \emph{search method} that follows this grammar. Therefore, each SLC algorithm has a probability of $\approx$3.571\% of being selected.

The second component of problem transformation methods is the actual problem transformation algorithm to deal with the single-label classification.
In other words, this step defines the choice of the MLC algorithm to handle the results created by the single-label classification models.
For this component,  we divided its respective algorithms into three categories, i.e., three production rules in the grammar (see Figure \ref{fig_gramar5}): 
\text{\textless ALGS-PT1\textgreater}, \text{\textless ALGS-PT2\textgreater} and  \text{\textless ALGS-PT3\textgreater}.
The main reason for the creation of these (sub-)categories is related to the constraints of the multi-label meta-algorithms in the MEKA software. 
Although all MLC algorithms can be used in a standalone fashion, 
they can also be combined with multi-label meta-algorithms.
In MEKA, some MLC algorithms work very well at the multi-label base level of meta-algorithms, whereas others do not.
Thus, we had to create rules in the grammar to overcome the limitations in the used software.
The next paragraphs will refer to the Figures \ref{fig_gramar5}~and~\ref{fig_gramar7} to explain these links and constraints 
between problem transformation algorithms and multi-label meta-algorithms.

\begin{figure*}[!ht]
\fontsize{7pt}{8pt}\selectfont
\centering    
\begin{Verbatim}[frame=single,samepage=true] 
<ALGS-PT> ::= <ALGS-PT1> | <ALGS-PT2> | <ALGS-PT3>  
                                                            
                                                            
<ALGS-PT1> ::= BR | CC | LC | (BRq | CCq) <dsr> |           #BR=`Binary Relevance', CC=`Classifier Chain'
               <ComplexCC_Trellis> | FW | RT | <LP_based>   #LC=`Label Combination'
							    #BRq and CCq = `quick versions for BR and CC'
                                                            #FW=`Four-class pairWise', RT=`Ranking-Threshold'
               
               
<ALGS-PT2> ::= BCC <dp_complete>                            #BCC=`Bayesian Classifier Chain'
<ALGS-PT3> ::= PMCC <B> <ts> <ii> <chi_PMCC> <ps> <pof>     #PMCC=`Population of Monte-Carlo Classifier Chains'

<dsr> ::= RANDOM-REAL(0.2, 0.8)                             #dsr=`down-sample ratio'

<ComplexCC_Trellis> ::= PCC | (MCC <chi_MCC> | <CT>) <ii> <pof> |             
                       (CDN | <CDT>) <i_cdn_cdt> <ci>       #PCC=`Probabilistic Classifier Chains'
                                                            #MCC=`Monte-Carlo Classifier Chains'
                                                            #CT=`Classication Trellis'
                                                            #CDN=`Conditional Dependency Networks'
                                                            #CDT=`Conditional Dependency Trellis'                                        

<chi_MCC>::= <chi_CT> | 0                                   #chi_MCC=`nmber of chain iterations for MCC'
<ii> ::= RANDOM-INT(2, 100)                                 #ii=`number of inference interations'
<pof> ::= Accuracy | Jaccard index | Hamming score | Exact match | Jaccard distance | Rank loss |
          Hamming loss | Zero One loss | Harmonic score | Log Loss lim:L | Micro Recall | One error | 
          Log Loss lim:D | Micro Precision | Macro Precision | Macro Recall | F1 micro averaged | 
          Avg precision | F1 macro averaged by example | F1 macro averaged by label | AUPRC macro averaged |  
          AUROC macro averaged | Levenshtein distance 
							    #pof=`Payoff function'

<CT> ::= <chi_CT> <w> <dp>
<dp> ::= C | I | Ib | Ibf | H | Hbf | X | F | None          #dp=`dependency type'
<chi_CT> ::= RANDOM-INT(2, 1500)                            #chi_CT=`number of chain iterations for CT'
<w> ::= 0 1 | -1 <d>                                        #w=`width of the trellis'
<d> ::= RANDOM-INT(1, SQRT(L) +1)                           #d=`neighborhood density'
							    #Where L is the number of labels
<CDT> ::= <w> <dp>                                          #parameters defined earlier

<i_cdn_cdt> ::= RANDOM-INT(101, 1000)                       #i_cdn_cdt=`total number of iterations'
<ci> ::= RANDOM-INT(1, 100)                                 #ci=`collection iterations'

<LP_based> ::= (PS | PSt | <RAkEL-based> ) <sv> <pv>        #PS=`Pruned Sets'
                                                            #PSt=`Pruned Sets with Threshold'
<sv> ::= RANDOM-INT(0, 5)                                   #sv=`subsampling value'                                          
<pv> ::= RANDOM-INT(1, 5)                                   #pv=`pruning value'
 
<RAkEL-based> ::= (RAkEL <sre> | RAkELd)  <les>             #RAkEL=`RAndom k-labEL Pruned Sets'
                                                            #RAkELd=`RAndom k-labEL Disjoint Pruned Sets'
<sre> ::= RANDOM-INT(2, min(2L, 100) )                      #sre=`number of subsets to run in an ensemble'
<les> ::= RANDOM-INT(1, L/2)                                #les=`number of labels in each label subset'
							    #Where L is the number of labels
<dp_complete> ::= <dp> | LEAD                               #dp=`complete dependency type for BCC'                  

<B> ::= RANDOM-REAL(0.01, 0.99)                             #B=`Beta factor for deacreasing the temperature'
<ts> ::= 0 | 1                                              #ts=`Temperature switch'
<ps> ::=  RANDOM-INT(1, 50)                                 #ps=`population size'
<chi_PMCC> ::= RANDOM-INT(51, 1500)                         #chi_PMCC=`number of chain iterations for PMCC'
\end{Verbatim}
    \setlength{\fboxsep}{1mm}
    \caption{Defined Grammar -- Part 5: MLC Problem Transformation Methods.}
    \label{fig_gramar5}
\end{figure*}


We referred to the first production rule to define problem transformation methods as 
\text{\textless ALGS-PT1\textgreater} in Figure~\ref{fig_gramar5}. This rule encompasses the traditional algorithms BR, CC and LC (which is also known as LP). Besides, it includes the quick versions of BR and CC (i.e, BRq and CCq),
all the complex classifier chains and trellis algorithms (which are defined by the rule \text{\textless ComplexCC\_Trellis\textgreater}),
Four-class pairWise (FW), Ranking and Threshold (RT), and all the label powerset based algorithms (which are defined by the rule \text{\textless LP\_based\textgreater}).
The production rule \text{\textless ALGS-PT1\textgreater} is presented in the following rules in Figure~\ref{fig_gramar7}: \text{\textless META-MLC1\textgreater} (via \text{\textless ALGS-PT\textgreater}), \text{\textless META-MLC2\textgreater} and \text{\textless META-MLC3\textgreater}.
This means that this category of PT methods describes the majority of the MLC algorithms in MEKA (84.21\% of the cases, i.e., 16 of the 19 MLC algorithms) and, 
in addition, all these algorithms can be combined with all meta-algorithms in the MEKA software. Thus,
\text{\textless ALGS-PT1\textgreater} can be considered the least restrictive  of the PT method rules in the grammar.

\text{\textless ALGS-PT2\textgreater}, in Figure~\ref{fig_gramar5}, is the production rule to describe solely the Bayesian Classifier Chain (BCC) algorithm, one of the most constrained algorithms
in the MEKA software. The BCC algorithm can only be executed in a standalone fashion or combined with the algorithms described by the production rules \text{\textless META-MLC1\textgreater} (via \text{\textless ALGS-PT\textgreater})
and \text{\textless META-MLC3\textgreater}. This means that BCC can be used with four (4) of the seven (7) meta-algorithms (in Figure~\ref{fig_gramar7}): 
Subset Mapper (SM), Random Subspace Multi-Label (RSML), Expectation Maximization (EM) and 
Classification Maximization (CM). In other cases of trying to use BCC, this will result in errors in MEKA's output and, therefore, this was not allowed in the grammar.

Similarly to \text{\textless ALGS-PT2\textgreater}, we have \text{\textless ALGS-PT3\textgreater}, a problem transformation rule that represents 
the Population of Monte-Carlo Classifier Chains~(PMCC) algorithm. This algorithm can only be used by itself and at the multi-label base level of 
five (5) of the seven (7) meta-algorithms:
Subset Mapper (SM), Random Subspace Multi-Label (RSML), Bagging of Multi-Label methods (BaggingML), Bagging of Multi-Label methods with Duplicates (BaggingMLDup) and Ensemble of Multi-Label methods (EnsembleML).
These five multi-label meta-algorithms are defined by the production rules \text{\textless META-MLC1\textgreater} and \text{\textless META-MLC2\textgreater}.
Therefore, the creation of \text{\textless ALGS-PT2\textgreater} is justified by the fact that PMCC algorithm can only be combined with these meta-algorithms, i.e.,
a constraint that did not appear in the other rules of the grammar.


Besides the problem transformation methods, we also have a multi-label version of the back propagation algorithm for training neural networks, called ML-BPNN.
This algorithm can be seen in the Figure \ref{fig_gramar6} and is the only one (for now) representing the algorithm adaptation (AA) methods, 
defined by the production rule \text{\textless MLC-AA\textgreater}.
ML-BPNN can also be associated to meta-algorithms. As we can see in Figure \ref{fig_gramar7}, this MLC algorithm can be linked to the meta-algorithms defined by the 
production rules \text{\textless META-MLC1\textgreater}, \text{\textless META-MLC2\textgreater} and \text{\textless META-MLC3\textgreater}.

\begin{figure*}[!ht]
\fontsize{7pt}{8pt}\selectfont
\centering        
\begin{Verbatim}[frame=single,samepage=true]
<MLC-AA> ::= <ML-BPNN>

<ML-BPNN> ::= <ne> <nhu_bpnn> <lr_bpnn> <m_bpnn>                 	#ML-BPNN=`Multi-Label Back Propagation 
								        #         Neural Network'
<ne> ::= RANDOM-INT(10, 1000)                                           #ne=`number of epochs'
<nhu_bpnn> ::= RANDOM-REAL(0.2, 1.0) * n_attributes                     #nhu_bpnn=`number of hidden units, that
                                                                        #is a parameter that dependes on the 
                                                                        #number of attributes of the dataset'

<lr_bpnn> ::= RANDOM-REAL(0.001, 0.1)                                   #lr_bpnn=`learning rate for BPNN/DBPNN'
<m_bpnn> ::=  RANDOM-REAL(0.2, 0.8)                                     #m_bpnn=`momentum for BPNN and DBPNN'
\end{Verbatim}
    \setlength{\fboxsep}{1mm}
    \caption{Defined Grammar -- Part 6: MLC Algorithm Adaptation Methods.}
    \label{fig_gramar6}
\end{figure*}

Finally, Figure \ref{fig_gramar7} covers all the multi-label meta-algorithms, which are defined by the production rule \text{\textless META-MLC-LEVEL\textgreater }.
As we explained previously, we created the production rules \text{\textless META-MLC1\textgreater}, \text{\textless META-MLC2\textgreater} and
\text{\textless META-MLC3\textgreater} in order to expand these five rules into \text{\textless META-MLC-LEVEL\textgreater}  to control the limitations, constraints and dependencies of the MEKA software
between meta-algorithms and multi-label algorithms (problem transformation and algorithm adaptation methods).
%



\begin{figure*}[!ht]
\fontsize{7pt}{8pt}\selectfont
\centering    
\begin{Verbatim}[frame=single,samepage=true]
<META-MLC-LEVEL> ::= <META-MLC1> | <META-MLC2> | <META-MLC3>
							     #META-MLC 1-3=`meta MLC algorithms                                         
							     # with different constraints' 
							     
<META-MLC1> ::= (SM | <RSML>) (<ALGS-PT> <ALGS-SLC> | <ML-BPNN>)
							     #SM=`Subset Mapper -- MLC method as parameter'

<RSML> ::= <bsp> <i_metamlc> <ap>                            #RSML=`Random Subspace Multi-Label'
<bsp> ::= RANDOM-INT(10, 100)                                #bsp=`bag size percent'
<i_metamlc> ::= RANDOM-INT(10, 50)                           #i_metamlc=`number of iterations for 
							     #meta MLC methods'
<ap> ::= RANDOM-INT(10, 100)                                 #ap=`attribute percent'

<META-MLC2> ::=  <alg-meta-mlc2> (<ALGS-PT1> | <ALGS-PT3>)  <ALGS-SLC> | <ML-BPNN>)

<alg-meta-mlc2> ::= ((BaggingML | BaggingMLDup <bsp> ) | EnsembleML <bsp_ensembleML>) <i_metamlc>  
                                                             #BaggingML=`Bagging of Multi-Label methods'
                                                             #BaggingMLDup=`BaggingML with duplicates'
                                                             #EnsembleML=`Ensemble of Multi-Label methods'
                                                             #bsp=`bag size percent -- defined earlier'     
								
<bsp_ensembleML> ::= RANDOM-INT(52, 72)                      #bsp_ensembleML=`specific bsp for EnsembleML'

<META-MLC3> ::= ( (EM | CM ) <i_metamlc> ) (<ALGS-PT1> | <ALGS-PT2>) <ALGS-SLC> | <ML-BPNN>)  
                                                             #EM=`Expectation Maximization'
                                                             #CM=`Classification Maximization'                                                     
\end{Verbatim}
    \setlength{\fboxsep}{1mm}
    \caption{Defined Grammar -- Part 7: MLC Meta-Algorithms.}
    \label{fig_gramar7}
\end{figure*}

\bibliographystyle{ACM-Reference-Format}
\bibliography{bibfile.bib} 

\end{document}